\documentclass[lettersize,journal]{IEEEtran}
\usepackage{amsmath,amsfonts}
\usepackage{algorithmic}
\usepackage{algorithm}
\usepackage{array}
\usepackage[caption=false,font=normalsize,labelfont=sf,textfont=sf]{subfig}
\usepackage{textcomp}
\usepackage{stfloats}
\usepackage{url}
\usepackage{verbatim}
\usepackage{graphicx}
\usepackage{cite}
\graphicspath{figures}
\usepackage{tabularx}
\usepackage{booktabs}
\usepackage{dcolumn}
\usepackage{hyperref}
\usepackage{threeparttable}
\usepackage{multirow}
\usepackage{mathrsfs}
\begin{document}

\title{Video Demoireing using Focused-Defocused Dual-Camera System}

\author{Xuan Dong*, Xiangyuan Sun*, Xia Wang*, Jian Song, Ya Li, Weixin Li
        % <-this % stops a space
\thanks{* denotes equal contribution. Xuan Dong, Xiangyuan Sun, Xia Wang and Ya Li are with the School of Artificial Intelligence, Beijing University of Posts and Telecommunications. Jian Song is now the CTO of DeepScience Co., Ltd. Weixin Li is an associate professor with the School of Computer Science and Engineering, Beihang University. Correspondence: weixinli@buaa.edu.cn.} % <-this % stops a space
}

\markboth{Submitted to TPAMI}%
{Shell \MakeLowercase{\textit{et al.}}: A Sample Article Using IEEEtran.cls for IEEE Journals}

\maketitle

\begin{abstract}

Moire patterns, unwanted color artifacts in images and videos, arise from the interference between spatially high-frequency scene contents and the spatial discrete sampling of digital cameras. Existing demoireing methods primarily rely on single-camera image/video processing, which faces two critical challenges: 1) distinguishing moire patterns from visually similar real textures, and 2) preserving tonal consistency and temporal coherence while removing moire artifacts. To address these issues, we propose a dual-camera framework that captures synchronized videos of the same scene: one in focus (retaining high-quality textures but may exhibit moire patterns) and one defocused (with significantly reduced moire patterns but blurred textures). We use the defocused video to help distinguish moire patterns from real texture, so as to guide the demoireing of the focused video. We propose a frame-wise demoireing pipeline, which begins with an optical flow based alignment step to address any discrepancies in displacement and occlusion between the focused and defocused frames. Then, we leverage the aligned defocused frame to guide the demoireing of the focused frame using a multi-scale CNN and a multi-dimensional training loss. To maintain tonal and temporal consistency, our final step involves a joint bilateral filter to leverage the demoireing result from the CNN as the guide to filter the input focused frame to obtain the final output. Experimental results demonstrate that our proposed framework largely outperforms state-of-the-art image and video demoireing methods. Please visit our project page at \url{https://github.com/circle11111/dual_lens_demoireing}.

\end{abstract}

\begin{IEEEkeywords}
Video demoireing, focused-defocused dual camera.
\end{IEEEkeywords}

\section{Introduction}

\begin{figure}[t]
\begin{center}
   \includegraphics[width=\linewidth]{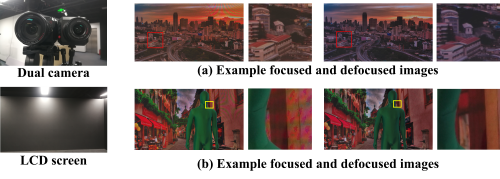}
\end{center}
   \caption{Left: The dual-camera system and LCD screen that are used in our experiment. Right: (a) and (b) give examples to show the problems of tonal inconsistency and relative displacement, respectively, between the focused and defocused frames. The proposed method addresses these problems and uses the defocused video as the guide to remove moire patterns in the focused video. }
\label{fig:inconsistency}
\end{figure}

\begin{figure*}[t]
\begin{center}
   \includegraphics[width=0.7\linewidth]{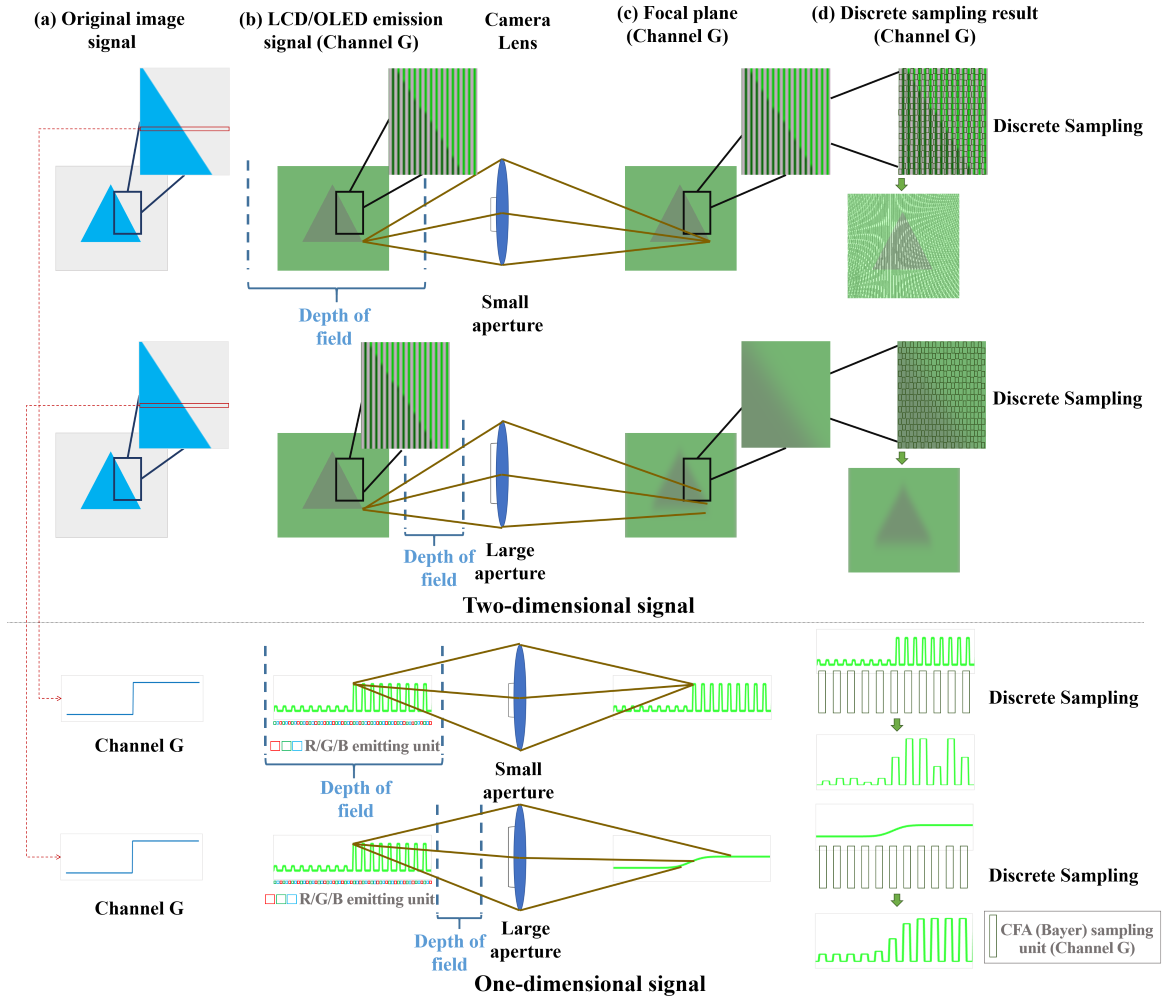}
\end{center}
   \caption{A toy example to explain the image formation and the generation of moire patterns in the focused and defocused frames. We only show the green (G) channel for simplification. In addition to the 2D image signal, the 1D signal of the pixels along a specific line of the image is demonstrated to highlight the intricate details of the image formation. (a) shows the original RGB image signal. (b) shows the emitted light of the green channel from the LCD/OLED screen. (c) shows the light that arrives at the focal plane. (d) shows the sampling results of the CMOS/CCD sensor at the focal plane with the color filter array (CFA) of Bayer mode.}
\label{fig:formation}
\end{figure*}

\begin{figure}[t]
\begin{center}
   \includegraphics[width=\linewidth]{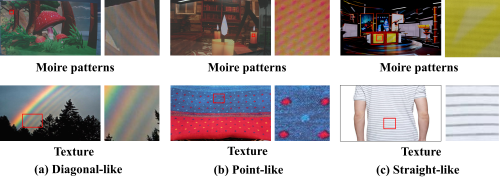}
\end{center}
   \caption{Example point-like and line-like moire patterns and textures to show the difficulty to differentiate moire patterns from real textures in real scenes.}
\label{fig:challenge}
\end{figure}

Liquid Crystal Display (LCD) and Organic Light-Emitting Diode (OLED) screens are widely utilized as display mediums for presenting information across various applications, \textit{e.g.} projecting PowerPoint slides at academic conferences, displaying product information in online shopping, providing visual background during live concerts, online weather forecasts, film production, \textit{etc}. These screens are usually shot together with the foreground presenters/objects into videos. However, in the shot videos, as the example images shown in Figs. \ref{fig:inconsistency}, \ref{fig:formation} and \ref{fig:challenge}, moire patterns frequently appear in the screen areas, even when high-end cameras are used. Consequently, video demoireing, \textit{i.e.} removing the moire patterns in the shot videos, has become a prevalent issue in both consumer and professional photography.

We show a toy example in Fig. \ref{fig:formation} to illustrate the mechanism behind the formation of moire patterns. 1) The light emitted from LCD/OLED screens is spatially periodic signal. In these screens, Red, Green, and Blue (RGB) emitting units are arranged side by side in a closely spaced pattern to produce a `full-color' composite signal. Due to this periodic arrangement, the intensity of light of each color channel, \textit{e.g.} green, changes spatially across the screen in a periodic manner. 2) In the discrete sampling of digital cameras, the RGB receiver units are positioned side by side as well, so the signal sampling of each color channel is also spatially periodic. 3) When the frequency of the spatially periodic signals emitted by the screen is comparable to the frequency of the camera's spatially discrete sampling, unwanted moire patterns may appear in the final shot video.

In the literature, researchers try to solve the demoireing problem by altering the camera lens hardware, \textit{e.g.} incorporating low-pass filters \cite{moire_18} \cite{moire_19}. 
While this can effectively diminish moire patterns, it simultaneously introduces blur and results in the loss of details. Another solution employs image/video processing methods to eliminate moire, \textit{e.g.} \cite{moire_14_42}. Most of the existing methods are designed for single-camera setups. However, due to the resemblance of moire patterns to repetitive and line-like real textures, as illustrated in Fig. \ref{fig:challenge}, these methods encounter difficulties in distinguishing moire from real textures. This challenge often results in persistent moire patterns and/or overly smoothed textures in the output.

Our insight is to set up a dual-camera system to capture two samples of the scene, as shown in Fig. \ref{fig:inconsistency}. The primary camera records a focused video, and the secondary camera records a defocused video. As Fig. \ref{fig:formation} explains, the focused video captures high-quality textures and may exhibit moire patterns. In the defocused video, the blur caused by defocusing smooths out fine details of textures, but meanwhile, the spatial period of the light signal from the screen is diminished and thus moire patterns are significantly reduced. 
Using the defocused video as guidance, we can robustly determine texture from moire patterns in the focused video, thus obtaining high-quality texture and removing moire patterns.

However, solving the demoireing problem using the dual-camera system is still a challenging task. As depicted in Fig. \ref{fig:inconsistency}, the challenges encompass 1) the tonal inconsistency between the focused video and the defocused video, 2) the presence of relative displacement and occlusions between the input pair due to disparity, 3) the need to remove moire patterns while preserving textures, and 4) ensuring that the results maintain temporal consistency with the input video. In this paper, an alignment component is incorporated into our pipeline to address the second challenge. We build a demoireing network with a multi-dimensional loss function to solve the third challenge. To tackle the first and fourth challenges, instead of directly utilizing the network's output as the final result, we employ it indirectly to assist in generating the final results. This is achieved by using the output of the demoireing network as guidance to filter the input focused frames, leveraging the filtering process's proficiency in maintaining tonal and temporal consistency.

\begin{figure*}[t]
\begin{center}
   \includegraphics[width=0.7\linewidth]{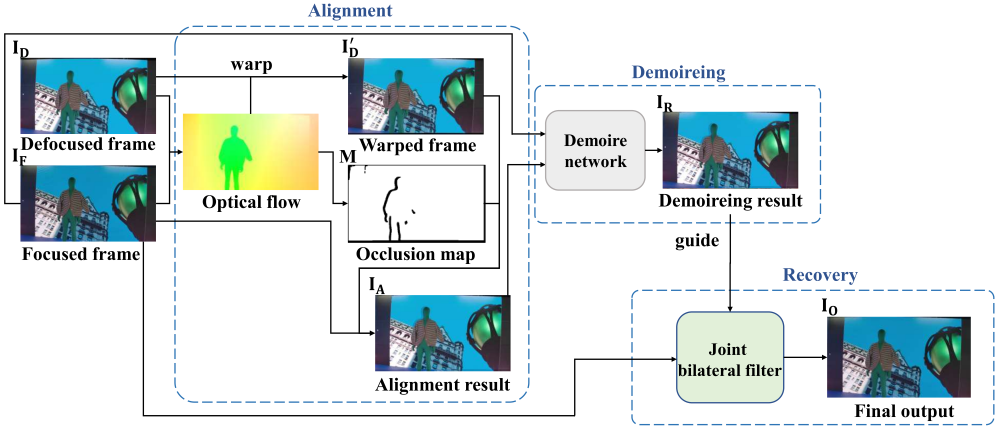}
\end{center}
   \caption{Our proposed frame-wise processing pipeline. In the \textbf{alignment} step, we warp the defocused frame $\bf{I_D}$ to align with the focused frame $\bf{I_F}$ and use the pixels of $\bf{I_F}$ to fill the occluded regions to obtain the alignment result $\bf{I_A}$. In the \textbf{demoireing} step, $\bf{I_F}$ is fed into the demoireing network to remove the moire patterns with $\bf{I_A}$ as guidance. The network is trained by a multi-dimensional loss. In the \textbf{recovery} step, $\bf{I_F}$ is filtered by the joint bilateral filter with $\bf{I_R}$ as guidance, to produce the final output $\bf{I_O}$ with tonal and temporal consistency.  }
\label{fig:pipeline}
\end{figure*}

Specifically, we propose a frame-wise processing pipeline for video demoireing, as outlined in Fig. \ref{fig:pipeline}. To address the relative displacement and occlusion challenges, the \textbf{alignment} step employs optical flow to align the defocused frame ${{\bf{I}}_{\bf{D}}}$ with the focused frame ${{\bf{I}}_{\bf{F}}}$, producing the warped result ${\bf{I}}_{\bf{D}}^{\bf{'}}$. The occluded area in ${\bf{I}}_{\bf{D}}^{\bf{'}}$ is replaced with pixels from ${{\bf{I}}_{\bf{F}}}$, to get the alignment result ${{\bf{I}}_{\bf{A}}}$. Then, in the \textbf{demoireing} step, we use ${{\bf{I}}_{\bf{F}}}$ and ${{\bf{I}}_{\bf{A}}}$ as inputs to the demoire network, yielding the moire-free result ${{\bf{I}}_{\bf{R}}}$. Here, ${{\bf{I}}_{\bf{A}}}$ guides moire removal in ${{\bf{I}}_{\bf{F}}}$. We design a novel multi-dimensional loss function for network training, including consistency, perceptual, high-frequency, and paired adversarial losses. Finally, to maintain tonal and temporal consistency, the \textbf{recovery} step utilizes ${{\bf{I}}_{\bf{R}}}$ as a guide to perform joint bilateral filtering on ${{\bf{I}}_{\bf{F}}}$, resulting in the final output ${{\bf{I}}_{\bf{O}}}$.

We generate three datasets for focused-defocused dual-camera demoireing, including one real dataset, named DualReal, one synthetic image dataset, named DualSynthetic, and one synthetic video dataset, named DualSyntheticVideo. Experiments are conducted on all datasets. The results of our proposed algorithm reveal considerable superiority over state-of-the-art image/video demoireing methods in moire removal. Moreover, our approach demonstrates the ability to maintain tonal and temporal consistency effectively.

Our \textbf{main contributions} are summarized as follows. \textbf{(1)} We propose a focused-defocused dual-camera setup for video demoireing, which effectively enhances the accuracy of moire removal without imposing restrictions on the user's camera. \textbf{(2)} We design a framewise processing framework for dual-camera video demoireing, which not only removes moire patterns by the demoireing network and multi-dimensional training losses, but also addresses challenges related to occlusions, tonal inconsistency and temporal flickering.

\begin{figure*}[t]
\begin{center}
   \includegraphics[width=0.8\linewidth]{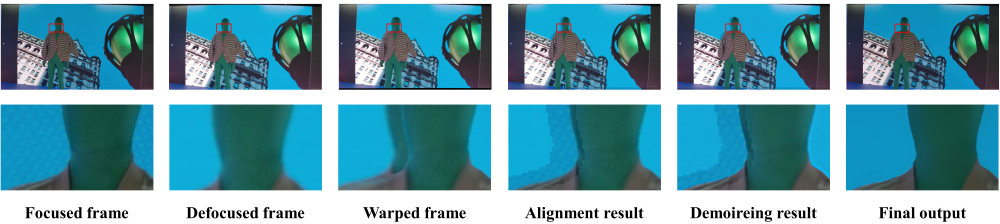}
\end{center}
   \caption{Intermediate results obtained through the pipeline. The regions marked in the red box are enlarged.}
\label{fig:inter_result}
\end{figure*}

\section{Related Work}

\subsection{Demoireing Methods}

Hardware solutions involve modifying the optical components of the camera to minimize spatially periodic changes in the light signal, thereby preventing moire patterns that result from the discrete sampling of CMOS/CCD. One example is the addition of an anti-aliasing low-pass filter \cite{moire_18} \cite{moire_19} in front of the camera lens. However, these methods encounter challenges in effectively eliminating moire while preserving image details, often leading to blurred images that compromise overall image quality.

The software solutions are primarily designed for the single-camera setup, encompassing single-image methods, multi-image methods, and video methods.

 In single-image methods, traditional algorithms include interpolation based algorithms \cite{moire_16} \cite{moire_17}, filter based algorithms \cite{moire_20} \cite{moire_24}, and layer decomposition based algorithms \cite{moire_21}. However, they struggle to remove large-scale moire patterns while preserving image texture. Deep learning based algorithms have proven to be more promising. The networks of the models include multi-scale networks such as those with UNet-like or Encoder-Decoder structures \cite{moire_2_27}\cite{moire_28}\cite{moire_29}\cite{moire_30}\cite{moire_31}\cite{moire_39}\cite{moire_6_36}\cite{moire_11_40}\cite{moire_5_35}, global-to-local networks, \textit{e.g.} \cite{3_70_33}, and transformed domain based networks, such as frequency \cite{moire_32} or wavelet domains \cite{moire_4_34}. Additionally, different loss functions are employed, including original or deformed L1 or L2 loss \cite{moire_2_27}\cite{moire_30}\cite{moire_32}, VGG for visual enhancement \cite{moire_4_34}\cite{moire_6_36}\cite{3_70_33}\cite{moire_38}, adversarial loss with prior knowledge from a discriminator \cite{moire_5_35}, and high-frequency domain loss \cite{moire_28}. 

  In multi-image demoireing algorithm, the work of \cite{moire_41} takes multiple frames as input and aligns them while using a multiscale feature encoding module to enhance low-frequency information. The work of \cite{moire_11_40} takes a focused image and a defocused image from the same camera and indirectly utilizes moire-free defocused images as a learning target in the training stage to help train the single-image based deep model for demoireing.
  
 In video demoireing algorithm \cite{moire_14_42}, adjacent frames are used to provide additional information for completing the mapping to moire-free images. This is achieved through multi-scale feature alignment, fusion, and de-moiring.

While existing methods are primarily tailored for single-camera videos/images, our paper introduces a novel video demoireing framework specifically designed for the focused-defocused dual-camera setup.

\subsection{Demoireing Dataset}

The demoireing datasets can be categorized into real datasets and synthetic datasets. 

Real datasets, such as \cite{moire_2_27}, \cite{moire_4_34}, \cite{moire_5_35}, \cite{3_70_33}, and \cite{moire_6_36}, are obtained by capturing moire-contaminated screen images/frames using a real camera and using a clean screenshot as the ground truth image/frame. However, these datasets suffer from differences in brightness and color between the ground truth and the moire image and limitations in the alignment accuracy. Additionally, foreground objects cannot be present in front of the screen, as they do not exist in the screenshot image. 

Synthetic datasets include single-image, such as \cite{moire_7}, \cite{moire_8}, \cite{moire_9}, and \cite{moire_10}, multi-image, such as \cite{transformer_44_13}, and video datasets, such as \cite{moire_14_42}. They simulate the formation of moire patterns to obtain synthetic moire images/frames, with the original image/frame without moire patterns serving as the ground truth image/frame. 

The work in \cite{moire_11_40} proposed both real and synthetic datasets consisting of pairs of focused moire images and defocused moire-free images from the same camera.

In this work, according to the proposed dual-camera setup, we introduce the focused-defocused dual-camera demoireing datasets, consisting of both synthetic and real datasets.

\subsection{Dual-camera Systems}

Dual-camera/multi-camera systems have been widely deployed in professional systems and smart phones. Many applications have been developed based on dual-camera systems, \textit{e.g.} super resolution \cite{Jeon18} \cite{WangL19}, colorization \cite{TIPDong22} \cite{dong2021self}, video retargeting \cite{Li18}, deblur \cite{Zhou19}, style transfer \cite{WangICCP19}, flow estimation \cite{Pan17}, and NeRF-based HDR \cite{Huang2022CVPR} \cite{Mildenhall2022CVPR}.

\section{Moire Formation in Focused-Defocused Images}

As shown in Fig. \ref{fig:inconsistency}, our system consists of two cameras placed side by side. This system offers flexibility to users, as it has no specific requirements for the primary camera, and users are free to choose their own cameras and imaging parameters. We impose certain constraints on the secondary camera. It has similar exposure levels with the primary camera. To shoot the defocused videos with reduced moire patterns, it has a large aperture to produce a shallow depth of field and is focused away from the LCD/OLED screen in the scene.

We present a toy example in Fig \ref{fig:formation}, to better demonstrate the formation of clear texture and moire patterns in the focused video, and the blurred texture and much less moire patterns in the defocused video. For the original signal (a), the emitted signal from the LCD/OLED screen is shown in (b). Due to the spatial side-by-side arrangement of red, green, and blue (RGB) emitting units, the light signal of any of the three colors, \textit{e.g.} green in Figure \ref{fig:formation}, which emits from the repetitive LCD/OLED emitting units, forms a periodically varying light signal that changes spatially across the screen. In the formation of the focused image, signal (b) is inside the depth of field (DoF), allowing the light emitted from any of its positions to propagate to the same position in the focal plane and thus be imaged clearly. However, in the formation of the defocused image, signal (b) is not within the DoF. According to the point spread function (PSF), the light emitted from any of the positions of signal (b) will propagate to different positions in space when imaged on the focal plane, resulting in defocused blur. This defocused blur not only blurs the texture of signal (a) but also reduces the apparent spatial variation of the light signal caused by the spatial side-by-side arrangement of RGB emitting units of the screen. This causes the light intensity at adjacent locations to tend to smooth out. When CCD/CMOS discrete sampling of the light signal at the focal plane is performed, the discrete sampled signal will show moire patterns in the focused image when the sampling frequency and the repetitive variation of the light signal are similar in frequency. In defocused images, the repetitive changes of the blurred optical signal are greatly reduced or disappear due to defocused blur, and therefore, the occurrence of moire patterns in the discrete sampled signal is greatly reduced.

\section{Method}

\subsection{Processing Pipeline}

We propose a frame-wise processing pipeline to mitigate the high computational costs associated with processing temporally neighboring multiple frames, with each frame treated similarly to image processing. Fig. \ref{fig:pipeline} shows the pipeline and Fig. \ref{fig:inter_result} shows the intermediate results of the key steps through our pipeline.

The alignment step utilizes the optical flow method of FlowFormer \cite{10.1007/978-3-031-19790-1_40}. We opted against employing the stereo matching algorithm due to its demanding requirements for dual-camera calibration and susceptibility to issues such as aberrations in the primary and secondary cameras, as well as relative movement resulting from shakes in practical applications. After calculating the optical flow ${\bf{F}}$, we warp the defocused frame ${{\bf{I}}_{\bf{D}}}$ captured by the secondary camera to obtain the warped result, \textit{i.e.} 
\begin{equation}
{\bf{I}}_{\bf{D}}^{\bf{'}} = warp({{\bf{I}}_{\bf{D}}},{\bf{F}}),
\end{equation}
and ${\bf{I}}_{\bf{D}}^{\bf{'}}$ aligns with the focus frame ${{\bf{I}}_{\bf{F}}}$ captured by the primary camera. We use the original focused data in ${{\bf{I}}_{\bf{F}}}$ to fill in the occluded regions of ${\bf{I}}_{\bf{D}}^{\bf{'}}$, resulting in the alignment result ${{\bf{I}}_{\bf{A}}}$, \textit{i.e.}
\begin{equation}
{{\bf{I}}_{\bf{A}}} = {\bf{I}}_{\bf{D}}^{\bf{'}} \cdot {\bf{M}} + {{\bf{I}}_{\bf{F}}} \cdot (1- {\bf{M}}),
\end{equation}
where $\bf{M}$ masks the occlusion regions in ${\bf{I}}_{\bf{D}}^{\bf{'}}$, which is computed by the standard forward-backward consistency check \cite{forward-backward_consistency_check}. While moire patterns may exist in the occluded region of ${{\bf{I}}_{\bf{A}}}$, this area is typically small owing to the limited baseline between the primary and secondary cameras, as depicted in Fig. \ref{fig:inconsistency}

In the demoireing step, we feed  ${{\bf{I}}_{\bf{A}}}$ and  ${{\bf{I}}_{\bf{F}}}$ into the demoireing network to learn the demoireing result ${{\bf{I}}_{\bf{R}}}$. We designed a multidimensional loss function to enhance the effectiveness of demoireing, including consistency, perceptual, high-frequency, and paired adversarial losses.

The demoireing result ${{\bf{I}}_{\bf{R}}}$ may suffer from tonal inconsistencies with the input frame and temporal flickering. To address this, in the recovery step, we use ${{\bf{I}}_{\bf{R}}}$ as the guide and perform the joint bilateral filter on ${{\bf{I}}_{\bf{F}}}$ to generate the final output frame ${{\bf{I}}_{\bf{O}}}$ by filtering ${{\bf{I}}_{\bf{F}}}$, \textit{i.e.}
\begin{equation}
\label{eqn:JBF}
{{\bf{I}}_{\bf{O}}} = JBF({{\bf{I}}_{\bf{F}}},{{\bf{I}}_{\bf{R}}}),
\end{equation}
where the joint bilateral filter $JBF$ is implemented by the fast bilateral filter of \cite{JBF_Bilat}. As the filtering is applied on ${{\bf{I}}_{\bf{F}}}$, leveraging the effectiveness of the bilateral filter in maintaining tonal and low-frequency consistency with the input frame, the final output frame ${{\bf{I}}_{\bf{O}}}$ exhibits better stability in both tonal and temporal consistency when compared to directly using ${{\bf{I}}_{\bf{R}}}$ as the final output. The moire is concurrently removed during the filtering process, as the filtering weight is computed based on the moire-free guide ${{\bf{I}}_{\bf{R}}}$.

\subsection{Demoireing Network}
\captionsetup[subfloat]{labelsep=none,format=plain,labelformat=empty}
\begin{figure*}[t]
\begin{center}
   \includegraphics[width=0.9\linewidth]{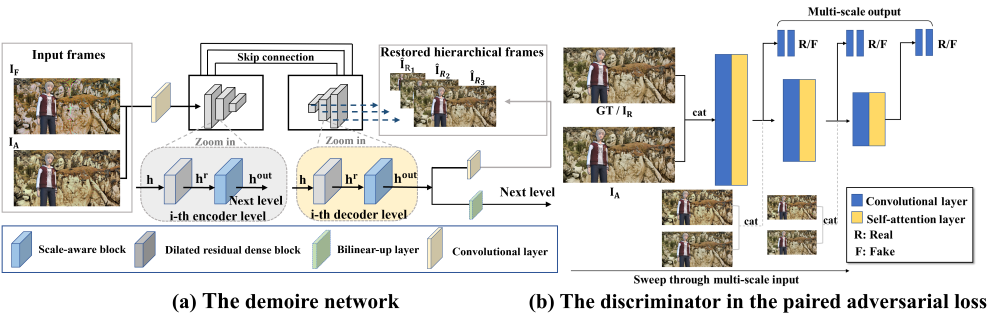}
\end{center}
   \caption{Network structures of (a) the demoireing network and (b) the discriminator in the paired adversarial loss.}
\label{fig:SAM}
\end{figure*}

As Fig. \ref{fig:SAM} (a) shows, the architecture of the demoireing network is based on the UNet structure. To enhance the network's performance, following \cite{moire_6_36}, we incorporate dilated residual dense blocks and scale aware module (SAM) blocks in each processing layer. To train the demoireing network, we employ synthetic data, as it provides ground-truth information for the synthetic input pair of focused and defocused images. This facilitates supervised learning, and the input pair of images is perfectly aligned. The multi-dimensional losses are listed below.

\textbf{Consistency loss.} To ensure that the output is consistent with the ground truth $\bf{GT}$, we employ the L1 loss, \textit{i.e.}
\begin{equation}
{L_{C}}{\rm{ = }}{\left\| {{\rm{{{\bf{I}}_{\bf{R}}} - {\bf{GT}}}}} \right\|_1},
\end{equation}
where ${{\bf{I}}_{\bf{R}}}$ is the output of the network and $\bf{GT}$ is the ground truth.

\textbf{Perceptual loss.} To improve the perceptual quality of the results, inspired by \cite{moire_6_36}, we use a pre-trained VGG16 \cite{vgg16} network, named $\Phi$, to extract features and design the multi-scale loss:

\begin{equation}
{L_P}{\rm{ = }}\sum\limits_{{\rm{i = 1}}}^3 {{{\left\| {\Phi \left( {{{\bf{I}}_{{{\bf{R}}_i}}}} \right){\rm{ - }}\Phi \left( {{\bf{G}}{{\bf{T}}_{\bf{i}}}} \right)} \right\|}_1}} ,
\end{equation}
where ${{{\bf{I}}_{{{\bf{R}}_i}}}}$ is the output in the $i$th level, and ${{{\bf{GT}}_{\bf{i}}}}$ is the corresponding ground-truth. Using the prior information of VGG, we are able to improve the subjective effect of the results, particularly in areas with distinct semantic features, such as sky and water.

\textbf{High-frequency loss.} To make better corrections to the high frequencies of the results, following \cite{FFT}, we use the high-frequency loss:
\begin{equation}
{L_{H}} = {\left\| {FFT({{\bf{I}}_{\bf{R}}}) - FFT({\bf{GT}})} \right\|_1},
\end{equation}
where $FFT( \bullet )$ represents the 2D Fast Fourier Transform. 

\captionsetup[subfloat]{labelsep=none,format=plain,labelformat=empty}
\begin{figure}[t]
\begin{center}
   \includegraphics[width=1\linewidth]{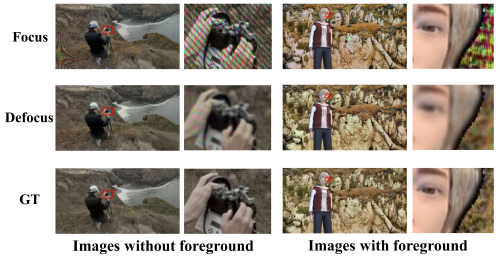}
\end{center}
   \caption{Example generated moire images in our synthetic dataset without and with foreground objects.}
\label{fig:synthetic_example}
\end{figure}

\textbf{Paired adversarial loss.} Besides the above manually designed losses, we design the discriminator $D$ to distinguish images with moire patterns from images without moire patterns. The network structure of the discriminator $D$ is shown in Fig.\ref{fig:SAM}. In this way, we can push the generator $G$, i.e. the demoire network, to learn to generate results $\bf{I_R}$ without moire patterns so as to fool $D$. The training of the generator $G$ and the training of the discriminator $D$ are conducted in an alternating manner.

When training $G$, we aim to generate $\bf{I_R}$ to fool $D$, and the paired adversarial loss ${L_{A}^G}$ is designed as  
\begin{equation}
{L_{A}^G}{\rm{ = }}\sum\limits_{{\rm{i = 1}}}^{{\rm{L}}} {\sum\limits_{{\rm{j = }}1}^{\rm{i}} {{{\left\| {{{{D}}_{{{ij}}}}(x, {\bf{I_A}}) - {{{V}}_{ij}^G}}(x) \right\|}_1}} },
\end{equation}
where $x$ is always $\bf{I_R}$, and the label of $x$, i.e. ${{V}}_{ij}^G(x)$, is real. Different from the traditional discriminator design, we also feed $\bf{I_A}$ into $D$ to construct the paired input $(x,{\bf{I_A}})$ of the paired adversarial loss, because $\bf{I_A}$ have blurred contents without moire patterns and can provide useful guidance to help $D$ judge whether $x$ contain moire patterns. Inspired by the work of GigaGAN \cite{gigagan}, we incorporated the multiscale-input and multiscale-output adversarial loss, and $i$, $j$ denote the level of the input scale and output scale, respectively. $L$ is the number of total levels.

When training $D$, we feed $\bf{I_R}$ or $\bf{GT}$ into $D$ , i.e. $x=\bf{I_R}$ or $x=\bf{GT}$, to let it learn to distinguish them. The corresponding loss $L_A^D$ is designed as
\begin{equation}
{L_{A}^D}{\rm{ = }}\sum\limits_{{\rm{i = 1}}}^{{\rm{L}}} {\sum\limits_{{\rm{j = }}1}^{\rm{i}} {{{\left\| {{{{D}}_{{{ij}}}}(x, {\bf{I_A}}) - {{V}}_{ij}^D}(x) \right\|}_1}} },
\end{equation}
 when $x=\bf{I_R}$, ${{V}}_{ij}^D(x)$ is fake, and when $x=\bf{GT}$, ${\bf{V}}_{ij}^D(x)$ is real. The values of real and fake are 1 and 0, respectively.

In training the generator $G$, i.e. the demoire network, as introduced above, multiple losses between $\bf{I_R}$ and $\bf{GT}$ are used, including consistency loss $L_C$, perceptual loss $L_P$, high-frequency loss $L_H$ and paired adversarial loss $L_A^G$, i.e. 
\begin{equation}
{L_{{\rm{total}}}} = {L_{C}} + {\lambda _P}{L_{P}} + {\lambda _H}{L_{H}} + {\lambda _A}{L_{A}^G},
\end{equation}
where $\lambda_{P}$, $\lambda_{H}$, and $\lambda_{A}$ are hyperparameters that control the relative importance. The values of $\lambda_{P}$, $\lambda_{H}$, and $\lambda_{A}$ are 1, 0.1 and 0.1 in this paper. 

\subsection{Synthetic Dataset}
\label{sec:synthetic}
The real dual-camera video data lacks the ground-truth video of the input focused video necessary for training deep models in our pipeline. Given our frame-wise processing strategy for video demoireing, we address this limitation by synthesizing an image dataset for supervised learning. This involves adding moire patterns into the original image at various levels to generate the input focused and defocused images, following the principles of moire generation \cite{moire_8}. The original image serves as the ground-truth image. The moire generation process includes the following steps:

\begin{enumerate}
    \item Resample the original image into a mosaic of RGB subpixels (modeled as a 3x3 grid with [R, G, B; R, G, B; R, G, B]) to simulate the image displayed on the LCD.
    \item Apply a random projective transformation to the image to simulate different relative positions and orientations of the display LCD and the camera. 
    \item Apply Gaussian blur when simulating the defocused camera, where the sigma value of the Gaussian filter is a random value between 3.2 and 4.0.
    \item Resample the image using the Bayer CFA to simulate the RAW data.
    \item Apply the demosaic function provided by MATLAB to convert the RAW data into RGB images, and then scale the results to compensate for brightness changes during the processing.
    \item Add a foreground to the image (optional). When simulating the defocused camera, we also apply Gaussian blur to the foreground of the defocused image, where the sigma value of the Gaussian filter is a random value between 2.0 and 2.5.
\end{enumerate}
We generate 5700 sets of images in total, and 3000 sets of the 5700 sets have foreground objects. Some example images are shown in Fig. \ref{fig:synthetic_example}. This synthetic image dataset is named DualSynthetic.

Besides DualSynthetic, we also build a synthetic focused-defocused dual-camera video demoireing dataset, named DualSyntheticVideo. The original video frames are from the famous stereo video dataset of Sintel\cite{sintel}. Sintel consists of 23 stereo videos of left view and right view, and each video contains 20-50 frames. In each video, the generation of the first pair of frames is performed according to the steps 1)-6) above. We add moire patterns to the left view frames and add blur to the right view frames to generate the input focused and defocused frames, respectively. The left view without adding moire patterns are used as ground-truth. The following frames are assumed with small and stable camera motion of the previous frame instead of random motion, and thus the step 2), i.e. the projective transformation, is not randomly generated. We fix the screen not moved and set a constant translation of the camera for each frame. For each video, the constant translation value is randomly selected between [5,20].

DualSynthetic contains plenty of images and types of moire patterns. The pairs of inputs in DualSyntheticVideo have mis-alignment and occlusions, which are more close to the real data captured by dual-camera systems.

\begin{figure*}
	\begin{minipage}[ht]{.99\linewidth}
		\centering
            \vspace{-0pt}
		\subfloat{\label{}\includegraphics[width=0.18\linewidth]{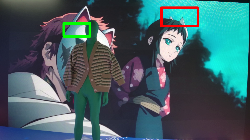}}\hspace{4pt}
		\subfloat{\label{}\includegraphics[width=0.18\linewidth]{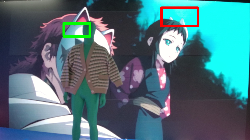}}\hspace{4pt}
		\subfloat{\label{}\includegraphics[width=0.18\linewidth]{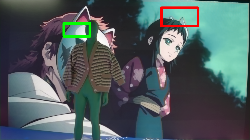}}\hspace{4pt}
		\subfloat{\label{}\includegraphics[width=0.18\linewidth]{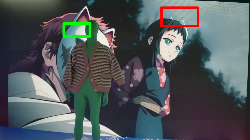}}\hspace{4pt}
		\subfloat{\label{}\includegraphics[width=0.18\linewidth]{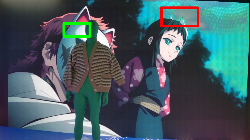}}\hspace{4pt}
	\end{minipage}
	\begin{minipage}[ht]{.99\linewidth}
		\centering
            \vspace{-7pt}
		\subfloat{\label{}\includegraphics[width=0.18\linewidth]{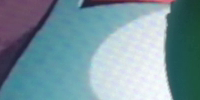}}\hspace{4pt}
		\subfloat{\label{}\includegraphics[width=0.18\linewidth]{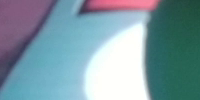}}\hspace{4pt}
		\subfloat{\label{}\includegraphics[width=0.18\linewidth]{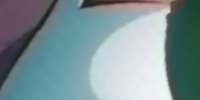}}\hspace{4pt}
		\subfloat{\label{}\includegraphics[width=0.18\linewidth]{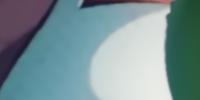}}\hspace{4pt}
		\subfloat{\label{}\includegraphics[width=0.18\linewidth]{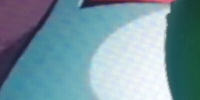}}\hspace{4pt}
	\end{minipage}
	\begin{minipage}[ht]{.99\linewidth}
		\centering
            \vspace{-7pt}
		\subfloat[\fontsize{8pt}{12pt}\selectfont Focused]{\label{}\includegraphics[width=0.18\linewidth]{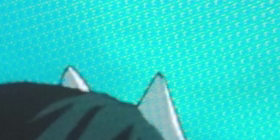}}\hspace{4pt}
		\subfloat[\fontsize{8pt}{12pt}\selectfont Defocused]{\label{}\includegraphics[width=0.18\linewidth]{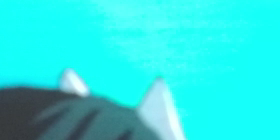}}\hspace{4pt}
		\subfloat[\fontsize{8pt}{12pt}\selectfont DMCNN]{\label{}\includegraphics[width=0.18\linewidth]{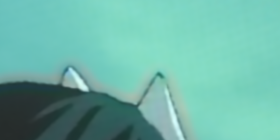}}\hspace{4pt}
		\subfloat[\fontsize{8pt}{12pt}\selectfont MopNet]{\label{}\includegraphics[width=0.18\linewidth]{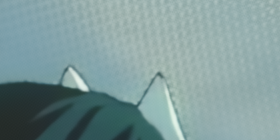}}\hspace{4pt}
		\subfloat[\fontsize{8pt}{12pt}\selectfont MBCNN]{\label{}\includegraphics[width=0.18\linewidth]{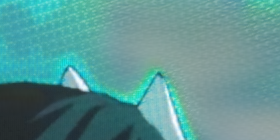}}\hspace{4pt}
	\end{minipage}
 %\vspace{-140pt}
	\begin{minipage}[ht]{.99\linewidth}
		\centering
  \vspace{-0pt}
		\subfloat{\label{}\includegraphics[width=0.18\linewidth]{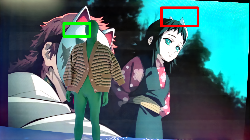}}\hspace{4pt}
		\subfloat{\label{}\includegraphics[width=0.18\linewidth]{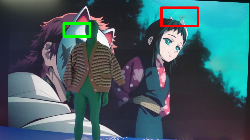}}\hspace{4pt}
		\subfloat{\label{}\includegraphics[width=0.18\linewidth]{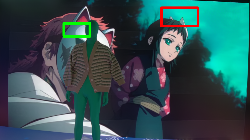}}\hspace{4pt}
		\subfloat{\label{}\includegraphics[width=0.18\linewidth]{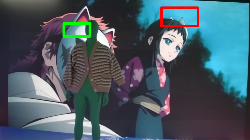}}\hspace{4pt}
		\subfloat{\label{}\includegraphics[width=0.18\linewidth]{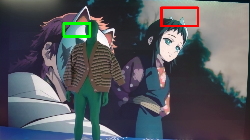}}\hspace{4pt}
	\end{minipage}
	\begin{minipage}[ht]{.99\linewidth}
		\centering
\vspace{-7pt}
		\subfloat{\label{}\includegraphics[width=0.18\linewidth]{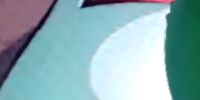}}\hspace{4pt}
		\subfloat{\label{}\includegraphics[width=0.18\linewidth]{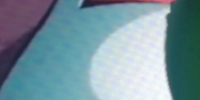}}\hspace{4pt}
		\subfloat{\label{}\includegraphics[width=0.18\linewidth]{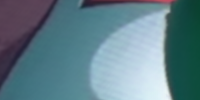}}\hspace{4pt}
		\subfloat{\label{}\includegraphics[width=0.18\linewidth]{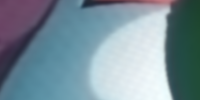}}\hspace{4pt}
		\subfloat{\label{}\includegraphics[width=0.18\linewidth]{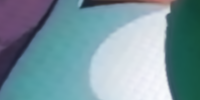}}\hspace{4pt}
	\end{minipage}
	\begin{minipage}[ht]{.99\linewidth}
		\centering
\vspace{-7pt}
		\subfloat[\fontsize{8pt}{12pt} \selectfont WDNet]{\label{}\includegraphics[width=0.18\linewidth]{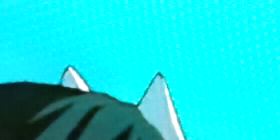}}\hspace{4pt}
		\subfloat[\fontsize{8pt}{12pt} \selectfont Uformer]{\label{}\includegraphics[width=0.18\linewidth]{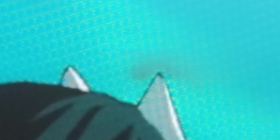}}\hspace{4pt}
		\subfloat[\fontsize{8pt}{12pt} \selectfont ESDNet]{\label{}\includegraphics[width=0.18\linewidth]{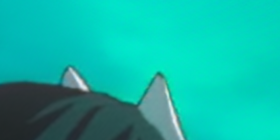}}\hspace{4pt}
		\subfloat[\fontsize{8pt}{12pt} \selectfont SwinIR]{\label{}\includegraphics[width=0.18\linewidth]{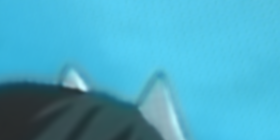}}\hspace{4pt}
		\subfloat[\fontsize{8pt}{12pt} \selectfont MFD]{\label{}\includegraphics[width=0.18\linewidth]{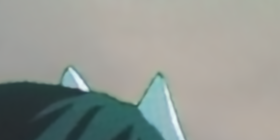}}\hspace{4pt}
	\end{minipage}
	%\vspace{-140pt}
	\begin{minipage}[ht]{.99\linewidth}
		\centering
\vspace{-0pt}
		\subfloat{\label{}\includegraphics[width=0.18\linewidth]{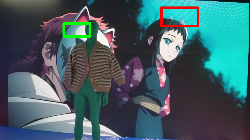}}\hspace{4pt}
		\subfloat{\label{}\includegraphics[width=0.18\linewidth]{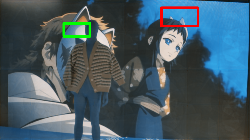}}\hspace{4pt}
		\subfloat{\label{}\includegraphics[width=0.18\linewidth]{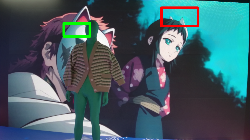}}\hspace{4pt}
		\subfloat{\label{}\includegraphics[width=0.18\linewidth]{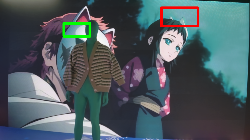}}\hspace{4pt}
		\subfloat{\label{}\includegraphics[width=0.18\linewidth]{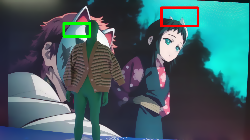}}\hspace{4pt}
	\end{minipage}
	\begin{minipage}[ht]{.99\linewidth}
		\centering
\vspace{-7pt}
		\subfloat{\label{}\includegraphics[width=0.18\linewidth]{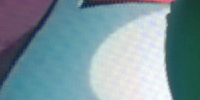}}\hspace{4pt}
		\subfloat{\label{}\includegraphics[width=0.18\linewidth]{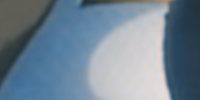}}\hspace{4pt}
		\subfloat{\label{}\includegraphics[width=0.18\linewidth]{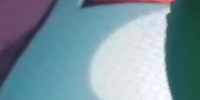}}\hspace{4pt}
		\subfloat{\label{}\includegraphics[width=0.18\linewidth]{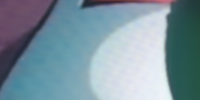}}\hspace{4pt}
		\subfloat{\label{}\includegraphics[width=0.18\linewidth]{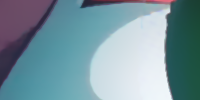}}\hspace{4pt}
	\end{minipage}
	\begin{minipage}[ht]{.99\linewidth}
		\centering
\vspace{-7pt}
		\subfloat[\fontsize{8pt}{12pt} \selectfont MMDM]{\label{}\includegraphics[width=0.18\linewidth]{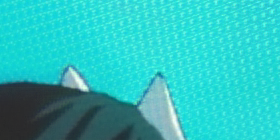}}\hspace{4pt}
		\subfloat[\fontsize{8pt}{12pt} \selectfont SiamTrans]{\label{}\includegraphics[width=0.18\linewidth]{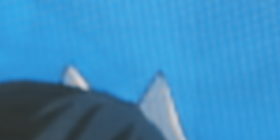}}\hspace{4pt}
		\subfloat[\fontsize{8pt}{12pt} \selectfont VDM$_{PCD}$]{\label{}\includegraphics[width=0.18\linewidth]{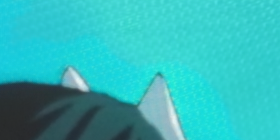}}\hspace{4pt}
		\subfloat[\fontsize{8pt}{12pt} \selectfont FDNet]{\label{}\includegraphics[width=0.18\linewidth]{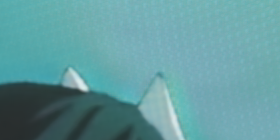}}\hspace{4pt}
		\subfloat[\fontsize{8pt}{12pt} \selectfont Ours]{\label{}\includegraphics[width=0.18\linewidth]{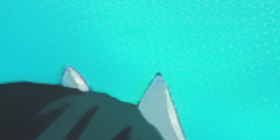}}\hspace{4pt}
	\end{minipage}
  \caption{Example results on DualReal. The regions marked with green and red boxes are enlarged. Please see more video and frame results in supplementary materials.}
 \label{fig:result_1}
\end{figure*}

\begin{figure*}
	\begin{minipage}[ht]{.99\linewidth}
		\centering
\vspace{-0pt}
		\subfloat{\label{}\includegraphics[width=0.18\linewidth]{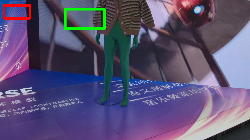}}\hspace{4pt}
		\subfloat{\label{}\includegraphics[width=0.18\linewidth]{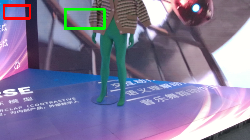}}\hspace{4pt}
		\subfloat{\label{}\includegraphics[width=0.18\linewidth]{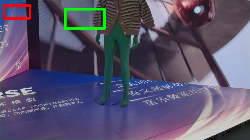}}\hspace{4pt}
		\subfloat{\label{}\includegraphics[width=0.18\linewidth]{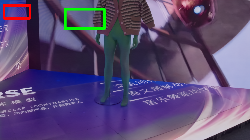}}\hspace{4pt}
		\subfloat{\label{}\includegraphics[width=0.18\linewidth]{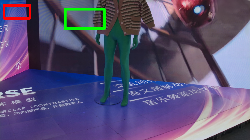}}\hspace{4pt}
	\end{minipage}
	\begin{minipage}[ht]{.99\linewidth}
		\centering
\vspace{-7pt}
		\subfloat{\label{}\includegraphics[width=0.18\linewidth]{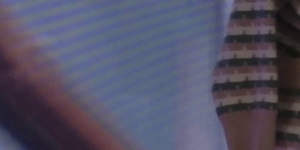}}\hspace{4pt}
		\subfloat{\label{}\includegraphics[width=0.18\linewidth]{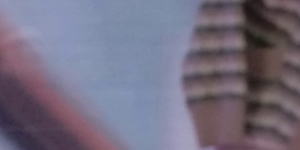}}\hspace{4pt}
		\subfloat{\label{}\includegraphics[width=0.18\linewidth]{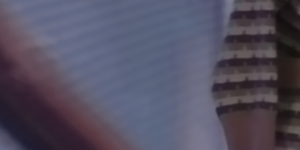}}\hspace{4pt}
		\subfloat{\label{}\includegraphics[width=0.18\linewidth]{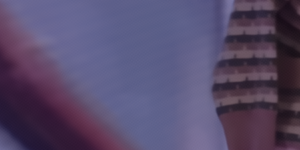}}\hspace{4pt}
		\subfloat{\label{}\includegraphics[width=0.18\linewidth]{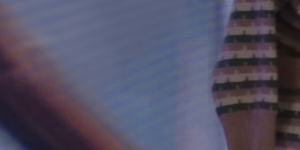}}\hspace{4pt}
	\end{minipage}
	\begin{minipage}[ht]{.99\linewidth}
		\centering
\vspace{-7pt}
		\subfloat[\fontsize{8pt}{12pt} \selectfont Focused]{\label{}\includegraphics[width=0.18\linewidth]{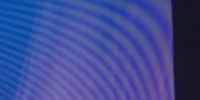}}\hspace{4pt}
		\subfloat[\fontsize{8pt}{12pt} \selectfont Defocused]{\label{}\includegraphics[width=0.18\linewidth]{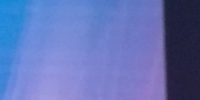}}\hspace{4pt}
		\subfloat[\fontsize{8pt}{12pt} \selectfont DMCNN]{\label{}\includegraphics[width=0.18\linewidth]{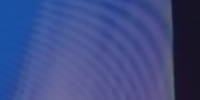}}\hspace{4pt}
		\subfloat[\fontsize{8pt}{12pt} \selectfont MopNet]{\label{}\includegraphics[width=0.18\linewidth]{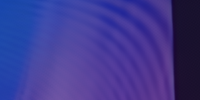}}\hspace{4pt}
		\subfloat[\fontsize{8pt}{12pt} \selectfont MBCNN]{\label{}\includegraphics[width=0.18\linewidth]{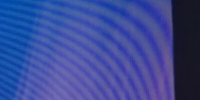}}\hspace{4pt}
	\end{minipage}
 %\vspace{-140pt}
	\begin{minipage}[ht]{.99\linewidth}
		\centering
\vspace{-0pt}
		\subfloat{\label{}\includegraphics[width=0.18\linewidth]{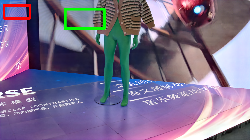}}\hspace{4pt}
		\subfloat{\label{}\includegraphics[width=0.18\linewidth]{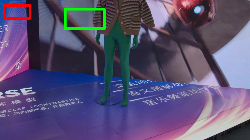}}\hspace{4pt}
		\subfloat{\label{}\includegraphics[width=0.18\linewidth]{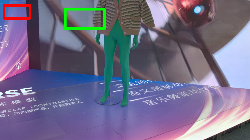}}\hspace{4pt}
		\subfloat{\label{}\includegraphics[width=0.18\linewidth]{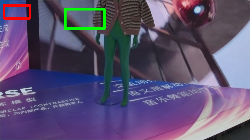}}\hspace{4pt}
		\subfloat{\label{}\includegraphics[width=0.18\linewidth]{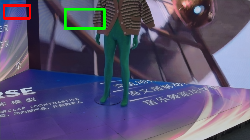}}\hspace{4pt}
	\end{minipage}
	\begin{minipage}[ht]{.99\linewidth}
		\centering
\vspace{-7pt}
		\subfloat{\label{}\includegraphics[width=0.18\linewidth]{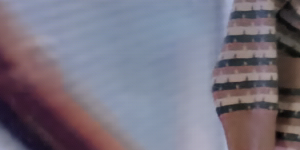}}\hspace{4pt}
		\subfloat{\label{}\includegraphics[width=0.18\linewidth]{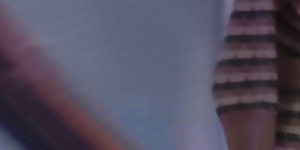}}\hspace{4pt}
		\subfloat{\label{}\includegraphics[width=0.18\linewidth]{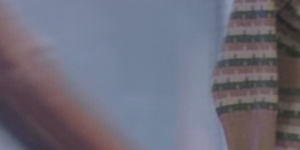}}\hspace{4pt}
		\subfloat{\label{}\includegraphics[width=0.18\linewidth]{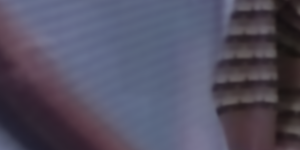}}\hspace{4pt}
		\subfloat{\label{}\includegraphics[width=0.18\linewidth]{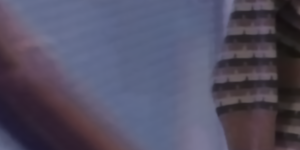}}\hspace{4pt}
	\end{minipage}
	\begin{minipage}[ht]{.99\linewidth}
		\centering
\vspace{-7pt}
		\subfloat[\fontsize{8pt}{12pt} \selectfont WDNet]{\label{}\includegraphics[width=0.18\linewidth]{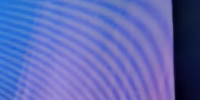}}\hspace{4pt}
		\subfloat[\fontsize{8pt}{12pt} \selectfont Uformer]{\label{}\includegraphics[width=0.18\linewidth]{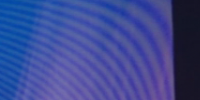}}\hspace{4pt}
		\subfloat[\fontsize{8pt}{12pt} \selectfont ESDNet]{\label{}\includegraphics[width=0.18\linewidth]{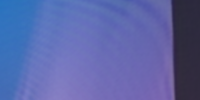}}\hspace{4pt}
		\subfloat[\fontsize{8pt}{12pt} \selectfont SwinIR]{\label{}\includegraphics[width=0.18\linewidth]{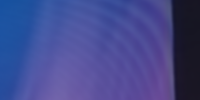}}\hspace{4pt}
		\subfloat[\fontsize{8pt}{12pt} \selectfont MFD]{\label{}\includegraphics[width=0.18\linewidth]{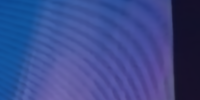}}\hspace{4pt}
	\end{minipage}
%\vspace{-140pt}
	\begin{minipage}[ht]{.99\linewidth}
		\centering
\vspace{-0pt}
		\subfloat{\label{}\includegraphics[width=0.18\linewidth]{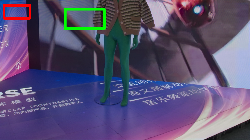}}\hspace{4pt}
		\subfloat{\label{}\includegraphics[width=0.18\linewidth]{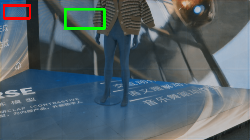}}\hspace{4pt}
		\subfloat{\label{}\includegraphics[width=0.18\linewidth]{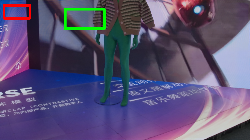}}\hspace{4pt}
		\subfloat{\label{}\includegraphics[width=0.18\linewidth]{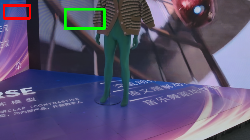}}\hspace{4pt}
		\subfloat{\label{}\includegraphics[width=0.18\linewidth]{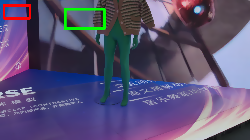}}\hspace{4pt}
	\end{minipage}
	\begin{minipage}[ht]{.99\linewidth}
		\centering
\vspace{-7pt}
		\subfloat{\label{}\includegraphics[width=0.18\linewidth]{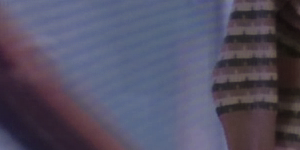}}\hspace{4pt}
		\subfloat{\label{}\includegraphics[width=0.18\linewidth]{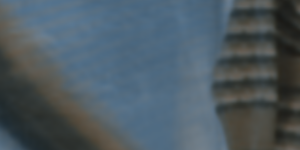}}\hspace{4pt}
		\subfloat{\label{}\includegraphics[width=0.18\linewidth]{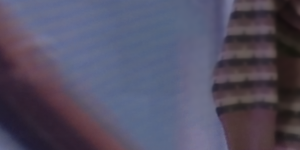}}\hspace{4pt}
		\subfloat{\label{}\includegraphics[width=0.18\linewidth]{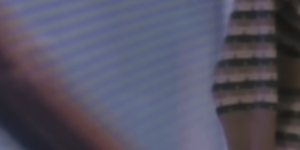}}\hspace{4pt}
		\subfloat{\label{}\includegraphics[width=0.18\linewidth]{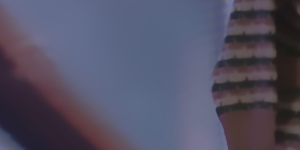}}\hspace{4pt}
	\end{minipage}
	\begin{minipage}[ht]{.99\linewidth}
		\centering
\vspace{-7pt}
		\subfloat[\fontsize{8pt}{12pt} \selectfont MMDM]{\label{}\includegraphics[width=0.18\linewidth]{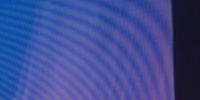}}\hspace{4pt}
		\subfloat[\fontsize{8pt}{12pt} \selectfont SiamTrans]{\label{}\includegraphics[width=0.18\linewidth]{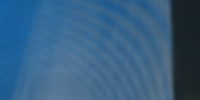}}\hspace{4pt}
		\subfloat[\fontsize{8pt}{12pt} \selectfont VDM$_{PCD}$]{\label{}\includegraphics[width=0.18\linewidth]{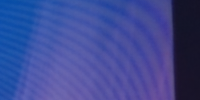}}\hspace{4pt}
		\subfloat[\fontsize{8pt}{12pt} \selectfont FDNet]{\label{}\includegraphics[width=0.18\linewidth]{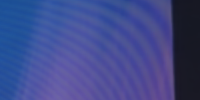}}\hspace{4pt}
		\subfloat[\fontsize{8pt}{12pt} \selectfont Ours]{\label{}\includegraphics[width=0.18\linewidth]{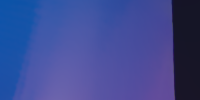}}\hspace{4pt}
	\end{minipage}
 
 \caption{Example results on DualReal. The regions marked with green and red boxes are enlarged. Please see more video and frame results in supplementary materials.}
 \label{fig:result_2}
\end{figure*}

\begin{figure*}
	\begin{minipage}[ht]{.99\linewidth}
		\centering
		\subfloat{\label{}\includegraphics[width=0.18\linewidth]{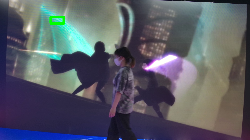}}\hspace{4pt}
		\subfloat{\label{}\includegraphics[width=0.18\linewidth]{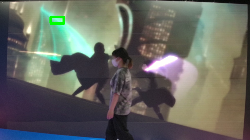}}\hspace{4pt}
		\subfloat{\label{}\includegraphics[width=0.18\linewidth]{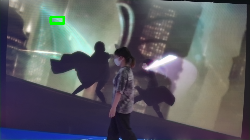}}\hspace{4pt}
		\subfloat{\label{}\includegraphics[width=0.18\linewidth]{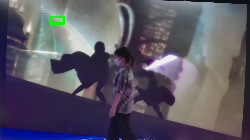}}\hspace{4pt}
		\subfloat{\label{}\includegraphics[width=0.18\linewidth]{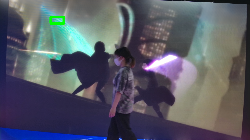}}\hspace{4pt}
	\end{minipage}
	\begin{minipage}[ht]{.99\linewidth}
		\centering
		\subfloat[\fontsize{8pt}{12pt} \selectfont Focused]{\label{}\includegraphics[width=0.18\linewidth]{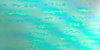}}\hspace{4pt}
		\subfloat[\fontsize{8pt}{12pt} \selectfont Defocused]{\label{}\includegraphics[width=0.18\linewidth]{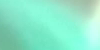}}\hspace{4pt}
		\subfloat[\fontsize{8pt}{12pt} \selectfont DMCNN]{\label{}\includegraphics[width=0.18\linewidth]{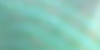}}\hspace{4pt}
		\subfloat[\fontsize{8pt}{12pt} \selectfont MopNet]{\label{}\includegraphics[width=0.18\linewidth]{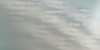}}\hspace{4pt}
		\subfloat[\fontsize{8pt}{12pt} \selectfont MBCNN]{\label{}\includegraphics[width=0.18\linewidth]{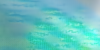}}\hspace{4pt}
	\end{minipage}

	%\vspace{-140pt}
	\begin{minipage}[ht]{.99\linewidth}
		\centering
		\subfloat{\label{}\includegraphics[width=0.18\linewidth]{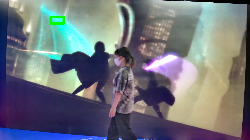}}\hspace{4pt}
		\subfloat{\label{}\includegraphics[width=0.18\linewidth]{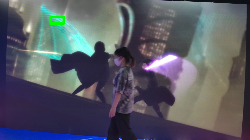}}\hspace{4pt}
		\subfloat{\label{}\includegraphics[width=0.18\linewidth]{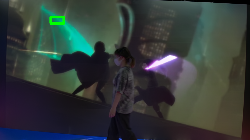}}\hspace{4pt}
		\subfloat{\label{}\includegraphics[width=0.18\linewidth]{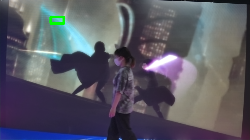}}\hspace{4pt}
		\subfloat{\label{}\includegraphics[width=0.18\linewidth]{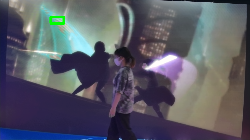}}\hspace{4pt}
	\end{minipage}
	\begin{minipage}[ht]{.99\linewidth}
		\centering
		\subfloat[\fontsize{8pt}{12pt} \selectfont WDNet]{\label{}\includegraphics[width=0.18\linewidth]{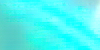}}\hspace{4pt}
		\subfloat[\fontsize{8pt}{12pt} \selectfont Uformer]{\label{}\includegraphics[width=0.18\linewidth]{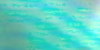}}\hspace{4pt}
		\subfloat[\fontsize{8pt}{12pt} \selectfont ESDNet]{\label{}\includegraphics[width=0.18\linewidth]{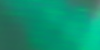}}\hspace{4pt}
		\subfloat[\fontsize{8pt}{12pt} \selectfont SwinIR]{\label{}\includegraphics[width=0.18\linewidth]{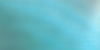}}\hspace{4pt}
		\subfloat[\fontsize{8pt}{12pt} \selectfont MFD]{\label{}\includegraphics[width=0.18\linewidth]{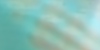}}\hspace{4pt}
	\end{minipage}

	%\vspace{-140pt}
	\begin{minipage}[ht]{.99\linewidth}
		\centering
		\subfloat{\label{}\includegraphics[width=0.18\linewidth]{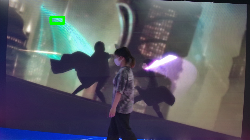}}\hspace{4pt}
		\subfloat{\label{}\includegraphics[width=0.18\linewidth]{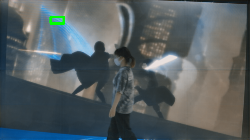}}\hspace{4pt}
		\subfloat{\label{}\includegraphics[width=0.18\linewidth]{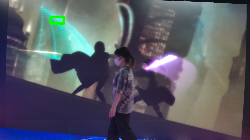}}\hspace{4pt}
		\subfloat{\label{}\includegraphics[width=0.18\linewidth]{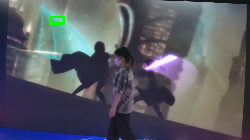}}\hspace{4pt}
		\subfloat{\label{}\includegraphics[width=0.18\linewidth]{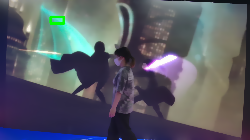}}\hspace{4pt}
	\end{minipage}
	\begin{minipage}[ht]{.99\linewidth}
		\centering
		\subfloat[\fontsize{8pt}{12pt} \selectfont MMDM]{\label{}\includegraphics[width=0.18\linewidth]{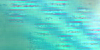}}\hspace{4pt}
		\subfloat[\fontsize{8pt}{12pt} \selectfont SiamTrans]{\label{}\includegraphics[width=0.18\linewidth]{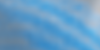}}\hspace{4pt}
		\subfloat[\fontsize{8pt}{12pt} \selectfont VDM$_{PCD}$]{\label{}\includegraphics[width=0.18\linewidth]{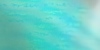}}\hspace{4pt}
		\subfloat[\fontsize{8pt}{12pt} \selectfont FDNet]{\label{}\includegraphics[width=0.18\linewidth]{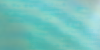}}\hspace{4pt}
		\subfloat[\fontsize{8pt}{12pt} \selectfont Ours]{\label{}\includegraphics[width=0.18\linewidth]{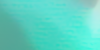}}\hspace{4pt}
	\end{minipage}
 \caption{Example results on DualReal. The regions marked with green boxes are enlarged. Please see more video and frame results in supplementary materials.}
\label{fig:result_3}
\end{figure*}

\begin{figure*}
	\begin{minipage}[ht]{.99\linewidth}
		\centering
		\subfloat{\label{}\includegraphics[width=0.18\linewidth]{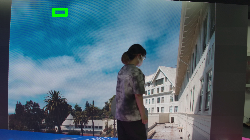}}\hspace{4pt}
		\subfloat{\label{}\includegraphics[width=0.18\linewidth]{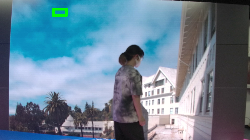}}\hspace{4pt}
		\subfloat{\label{}\includegraphics[width=0.18\linewidth]{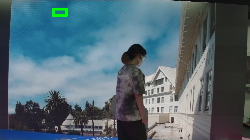}}\hspace{4pt}
		\subfloat{\label{}\includegraphics[width=0.18\linewidth]{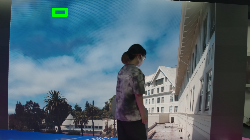}}\hspace{4pt}
		\subfloat{\label{}\includegraphics[width=0.18\linewidth]{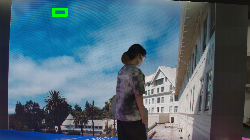}}\hspace{4pt}
	\end{minipage}
	\begin{minipage}[ht]{.99\linewidth}
		\centering
		\subfloat[\fontsize{8pt}{12pt} \selectfont Focused]{\label{}\includegraphics[width=0.18\linewidth]{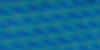}}\hspace{4pt}
		\subfloat[\fontsize{8pt}{12pt} \selectfont Defocused]{\label{}\includegraphics[width=0.18\linewidth]{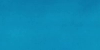}}\hspace{4pt}
		\subfloat[\fontsize{8pt}{12pt} \selectfont DMCNN]{\label{}\includegraphics[width=0.18\linewidth]{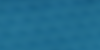}}\hspace{4pt}
		\subfloat[\fontsize{8pt}{12pt} \selectfont MopNet]{\label{}\includegraphics[width=0.18\linewidth]{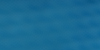}}\hspace{4pt}
		\subfloat[\fontsize{8pt}{12pt} \selectfont MBCNN]{\label{}\includegraphics[width=0.18\linewidth]{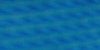}}\hspace{4pt}
	\end{minipage}

	%\vspace{-140pt}
	\begin{minipage}[ht]{.99\linewidth}
		\centering
		\subfloat{\label{}\includegraphics[width=0.18\linewidth]{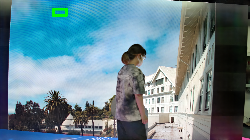}}\hspace{4pt}
		\subfloat{\label{}\includegraphics[width=0.18\linewidth]{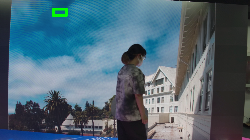}}\hspace{4pt}
		\subfloat{\label{}\includegraphics[width=0.18\linewidth]{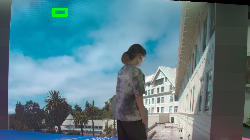}}\hspace{4pt}
		\subfloat{\label{}\includegraphics[width=0.18\linewidth]{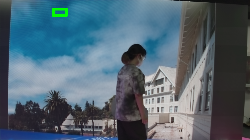}}\hspace{4pt}
		\subfloat{\label{}\includegraphics[width=0.18\linewidth]{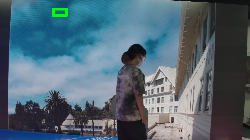}}\hspace{4pt}
	\end{minipage}
	\begin{minipage}[ht]{.99\linewidth}
		\centering
		\subfloat[\fontsize{8pt}{12pt} \selectfont WDNet]{\label{}\includegraphics[width=0.18\linewidth]{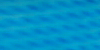}}\hspace{4pt}
		\subfloat[\fontsize{8pt}{12pt} \selectfont Uformer]{\label{}\includegraphics[width=0.18\linewidth]{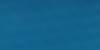}}\hspace{4pt}
		\subfloat[\fontsize{8pt}{12pt} \selectfont ESDNet]{\label{}\includegraphics[width=0.18\linewidth]{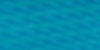}}\hspace{4pt}
		\subfloat[\fontsize{8pt}{12pt} \selectfont SwinIR]{\label{}\includegraphics[width=0.18\linewidth]{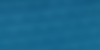}}\hspace{4pt}
		\subfloat[\fontsize{8pt}{12pt} \selectfont MFD]{\label{}\includegraphics[width=0.18\linewidth]{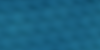}}\hspace{4pt}
	\end{minipage}

	%\vspace{-140pt}
	\begin{minipage}[ht]{.99\linewidth}
		\centering
		\subfloat{\label{}\includegraphics[width=0.18\linewidth]{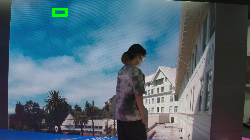}}\hspace{4pt}
		\subfloat{\label{}\includegraphics[width=0.18\linewidth]{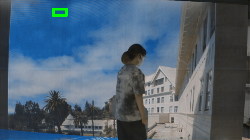}}\hspace{4pt}
		\subfloat{\label{}\includegraphics[width=0.18\linewidth]{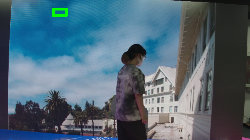}}\hspace{4pt}
		\subfloat{\label{}\includegraphics[width=0.18\linewidth]{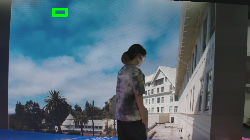}}\hspace{4pt}
		\subfloat{\label{}\includegraphics[width=0.18\linewidth]{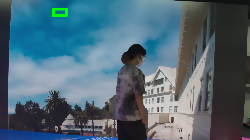}}\hspace{4pt}
	\end{minipage}
	\begin{minipage}[ht]{.99\linewidth}
		\centering
		\subfloat[\fontsize{8pt}{12pt} \selectfont MMDM]{\label{}\includegraphics[width=0.18\linewidth]{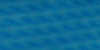}}\hspace{4pt}
		\subfloat[\fontsize{8pt}{12pt} \selectfont SiamTrans]{\label{}\includegraphics[width=0.18\linewidth]{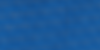}}\hspace{4pt}
		\subfloat[\fontsize{8pt}{12pt} \selectfont VDM$_{PCD}$]{\label{}\includegraphics[width=0.18\linewidth]{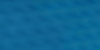}}\hspace{4pt}
		\subfloat[\fontsize{8pt}{12pt} \selectfont FDNet]{\label{}\includegraphics[width=0.18\linewidth]{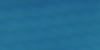}}\hspace{4pt}
		\subfloat[\fontsize{8pt}{12pt} \selectfont Ours]{\label{}\includegraphics[width=0.18\linewidth]{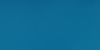}}\hspace{4pt}
	\end{minipage}
 \caption{Example results on DualReal. The regions marked with green boxes are enlarged. Please see more video and frame results in supplementary materials.}
\label{fig:result_4}
\end{figure*}

\begin{figure*}
\includegraphics[width=\linewidth]{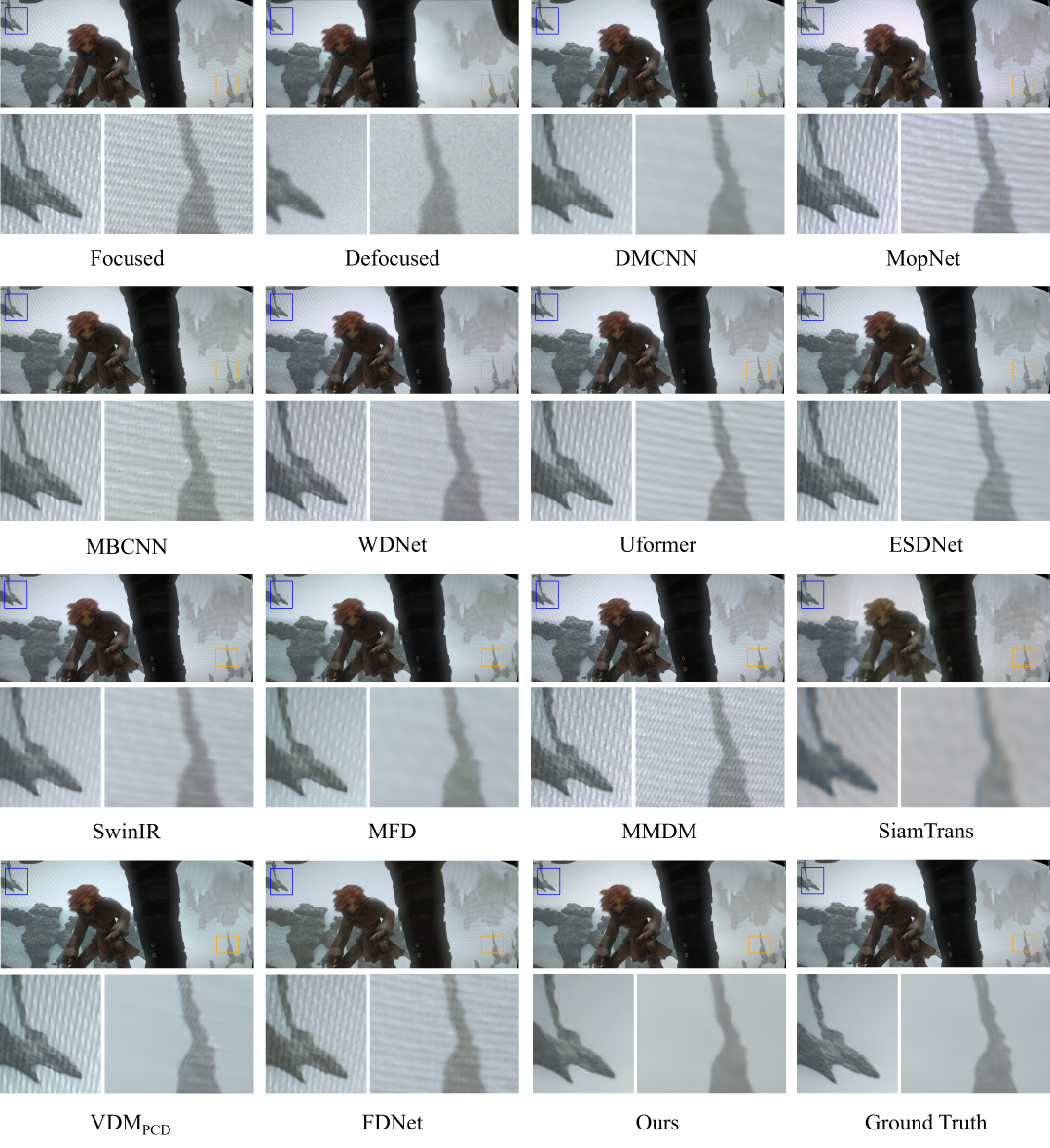}
  \caption{Example results on DualSyntheticVideo. The regions marked with green and red boxes are enlarged.}
 \label{fig:sintel_result_1}
\end{figure*}

\section{Experiment}
\captionsetup[subfloat]{labelsep=none,format=plain,labelformat=empty}
\begin{figure*}
	\vspace{-30pt}
	\begin{minipage}[ht]{.99\linewidth}
		\vspace{-0pt}
		\centering
		\subfloat{\label{}\includegraphics[width=0.21\linewidth]{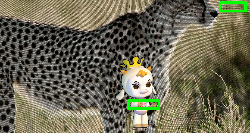}}\hspace{4pt}
		\subfloat{\label{}\includegraphics[width=0.21\linewidth]{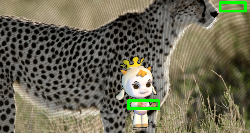}}\hspace{4pt}
		\subfloat{\label{}\includegraphics[width=0.21\linewidth]{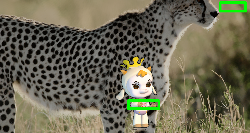}}\hspace{4pt}
		\subfloat{\label{}\includegraphics[width=0.21\linewidth]{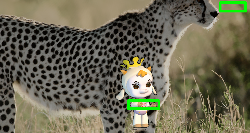}}\hspace{4pt}
	\end{minipage}
	\begin{minipage}[ht]{.99\linewidth}
		\vspace{-7pt}
		\centering
		\subfloat{\label{}\includegraphics[width=0.21\linewidth]{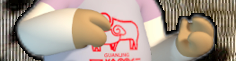}}\hspace{4pt}
		\subfloat{\label{}\includegraphics[width=0.21\linewidth]{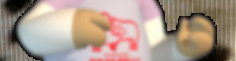}}\hspace{4pt}
		\subfloat{\label{}\includegraphics[width=0.21\linewidth]{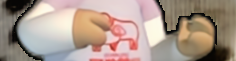}}\hspace{4pt}
		\subfloat{\label{}\includegraphics[width=0.21\linewidth]{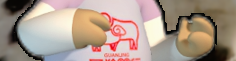}}\hspace{4pt}
	\end{minipage}
	\begin{minipage}[ht]{.99\linewidth}
		\vspace{-7pt}
		\centering
		\subfloat[\fontsize{8pt}{12pt} \selectfont Focused]{\label{}\includegraphics[width=0.21\linewidth]{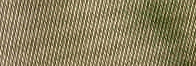}}\hspace{4pt}
		\subfloat[\fontsize{8pt}{12pt} \selectfont Defocused]{\label{}\includegraphics[width=0.21\linewidth]{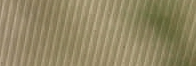}}\hspace{4pt}
		\subfloat[\fontsize{8pt}{12pt} \selectfont No foreground]{\label{}\includegraphics[width=0.21\linewidth]{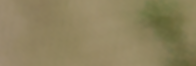}}\hspace{4pt}
		\subfloat[\fontsize{8pt}{12pt} \selectfont Ours]{\label{}\includegraphics[width=0.21\linewidth]{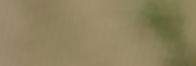}}\hspace{4pt}
	\end{minipage}
	\centering
	\fontsize{8pt}{12pt} \selectfont (a) Ablation of training data \\

	\begin{minipage}[ht]{.99\linewidth}
		\vspace{-0pt}
		\centering
		\subfloat{\label{}\includegraphics[width=0.13\linewidth]{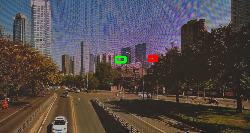}}\hspace{4pt}
		\subfloat{\label{}\includegraphics[width=0.13\linewidth]{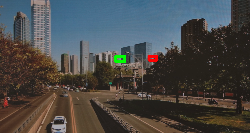}}\hspace{4pt}
		\subfloat{\label{}\includegraphics[width=0.13\linewidth]{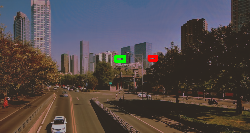}}\hspace{4pt}
		\subfloat{\label{}\includegraphics[width=0.13\linewidth]{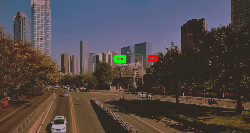}}\hspace{4pt}
		\subfloat{\label{}\includegraphics[width=0.13\linewidth]{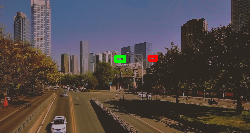}}\hspace{4pt}
		\subfloat{\label{}\includegraphics[width=0.13\linewidth]{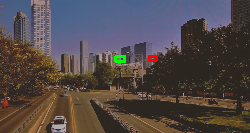}}\hspace{4pt}
		\subfloat{\label{}\includegraphics[width=0.13\linewidth]{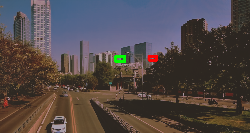}}\hspace{4pt}
	\end{minipage}
	\begin{minipage}[ht]{.99\linewidth}
		\vspace{-7pt}
		\centering
		\subfloat{\label{}\includegraphics[width=0.13\linewidth]{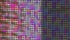}}\hspace{4pt}
		\subfloat{\label{}\includegraphics[width=0.13\linewidth]{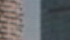}}\hspace{4pt}
		\subfloat{\label{}\includegraphics[width=0.13\linewidth]{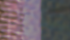}}\hspace{4pt}
		\subfloat{\label{}\includegraphics[width=0.13\linewidth]{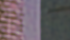}}\hspace{4pt}
		\subfloat{\label{}\includegraphics[width=0.13\linewidth]{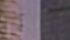}}\hspace{4pt}
		\subfloat{\label{}\includegraphics[width=0.13\linewidth]{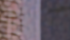}}\hspace{4pt}
		\subfloat{\label{}\includegraphics[width=0.13\linewidth]{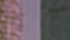}}\hspace{4pt}
	\end{minipage}
	\begin{minipage}[ht]{.99\linewidth}
		\vspace{-7pt}
		\centering
		\subfloat[\fontsize{8pt}{12pt} \selectfont Focused]{\label{}\includegraphics[width=0.13\linewidth]{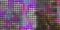}}\hspace{4pt}
		\subfloat[\fontsize{8pt}{12pt} \selectfont Defocused]{\label{}\includegraphics[width=0.13\linewidth]{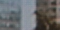}}\hspace{4pt}
		\subfloat[\fontsize{8pt}{12pt} \selectfont $L_C$]{\label{}\includegraphics[width=0.13\linewidth]{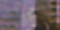}}\hspace{4pt}
		\subfloat[\fontsize{8pt}{12pt} \selectfont $L_C$+$L_P$]{\label{}\includegraphics[width=0.13\linewidth]{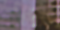}}\hspace{4pt}
		\subfloat[\fontsize{8pt}{12pt} \selectfont $L_C$+$L_P$+$L_H$]{\label{}\includegraphics[width=0.13\linewidth]{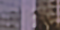}}\hspace{4pt}
		\subfloat[\fontsize{8pt}{12pt} \selectfont $L_C$+$L_P$+$L_A$]{\label{}\includegraphics[width=0.13\linewidth]{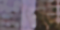}}\hspace{4pt}
		\subfloat[\fontsize{8pt}{12pt} \selectfont Ours]{\label{}\includegraphics[width=0.13\linewidth]{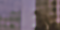}}\hspace{4pt}
	\end{minipage}
	\centering
	\fontsize{8pt}{12pt} \selectfont (b) Ablation of training loss \\
	
	\begin{minipage}[ht]{.99\linewidth}
		\vspace{-0pt}
		\centering
		\subfloat{\label{}\includegraphics[width=0.15\linewidth]{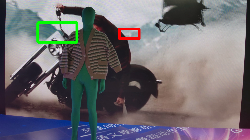}}\hspace{4pt}
		\subfloat{\label{}\includegraphics[width=0.15\linewidth]{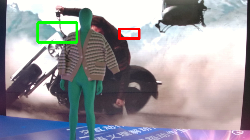}}\hspace{4pt}
		\subfloat{\label{}\includegraphics[width=0.15\linewidth]{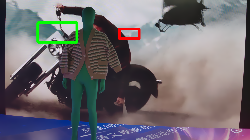}}\hspace{4pt}
		\subfloat{\label{}\includegraphics[width=0.15\linewidth]{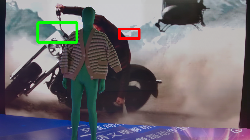}}\hspace{4pt}
		\subfloat{\label{}\includegraphics[width=0.15\linewidth]{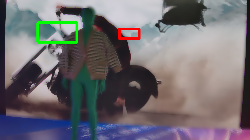}}\hspace{4pt}
		\subfloat{\label{}\includegraphics[width=0.15\linewidth]{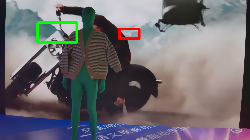}}\hspace{4pt}
	\end{minipage}
	\begin{minipage}[ht]{.99\linewidth}
		\vspace{-7pt}
		\centering
		\subfloat{\label{}\includegraphics[width=0.15\linewidth]{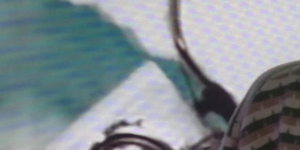}}\hspace{4pt}
		\subfloat{\label{}\includegraphics[width=0.15\linewidth]{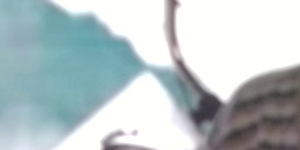}}\hspace{4pt}
		\subfloat{\label{}\includegraphics[width=0.15\linewidth]{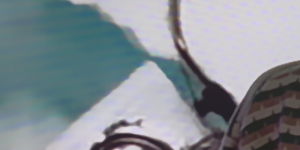}}\hspace{4pt}
		\subfloat{\label{}\includegraphics[width=0.15\linewidth]{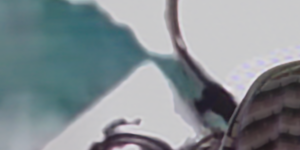}}\hspace{4pt}
		\subfloat{\label{}\includegraphics[width=0.15\linewidth]{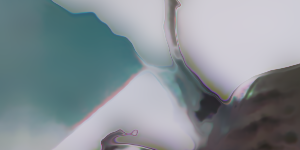}}\hspace{4pt}
		\subfloat{\label{}\includegraphics[width=0.15\linewidth]{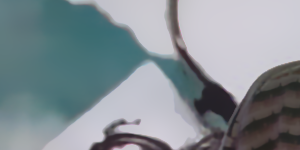}}\hspace{4pt}
	\end{minipage}
	\begin{minipage}[ht]{.99\linewidth}
		\vspace{-7pt}
		\centering
		\subfloat[\fontsize{8pt}{12pt} \selectfont Focused]{\label{}\includegraphics[width=0.15\linewidth]{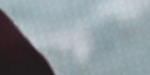}}\hspace{4pt}
		\subfloat[\fontsize{8pt}{12pt} \selectfont Defocused]{\label{}\includegraphics[width=0.15\linewidth]{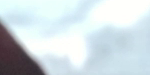}}\hspace{4pt}
		\subfloat[\fontsize{8pt}{12pt} \selectfont No defocus]{\label{}\includegraphics[width=0.15\linewidth]{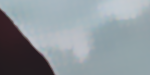}}\hspace{4pt}
		\subfloat[\fontsize{8pt}{12pt} \selectfont No JBF]{\label{}\includegraphics[width=0.15\linewidth]{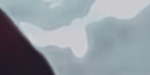}}\hspace{4pt}
		\subfloat[\fontsize{8pt}{12pt} \selectfont No Alignment]{\label{}\includegraphics[width=0.15\linewidth]{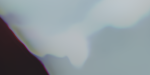}}\hspace{4pt}
		\subfloat[\fontsize{8pt}{12pt} \selectfont Ours]{\label{}\includegraphics[width=0.15\linewidth]{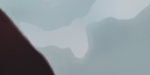}}\hspace{4pt}
	\end{minipage}
	\centering
	\fontsize{8pt}{12pt} \selectfont (c) Ablation of framework \\
	
	\begin{minipage}[ht]{.99\linewidth}
		\vspace{-0pt}
		\centering
		\subfloat{\label{}\includegraphics[width=0.15\linewidth]{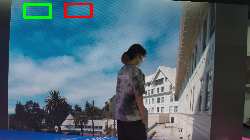}}\hspace{4pt}
		\subfloat{\label{}\includegraphics[width=0.15\linewidth]{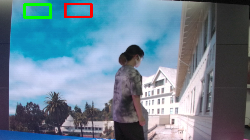}}\hspace{4pt}
		\subfloat{\label{}\includegraphics[width=0.15\linewidth]{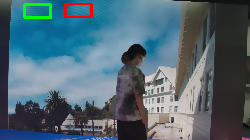}}\hspace{4pt}
		\subfloat{\label{}\includegraphics[width=0.15\linewidth]{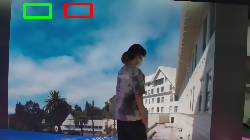}}\hspace{4pt}
		\subfloat{\label{}\includegraphics[width=0.15\linewidth]{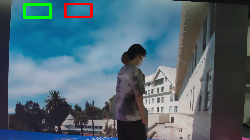}}\hspace{4pt}
		\subfloat{\label{}\includegraphics[width=0.15\linewidth]{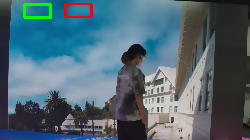}}\hspace{4pt}
	\end{minipage}
	\begin{minipage}[ht]{.99\linewidth}
		\vspace{-7pt}
		\centering
		\subfloat{\label{}\includegraphics[width=0.15\linewidth]{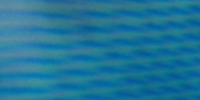}}\hspace{4pt}
		\subfloat{\label{}\includegraphics[width=0.15\linewidth]{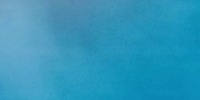}}\hspace{4pt}
		\subfloat{\label{}\includegraphics[width=0.15\linewidth]{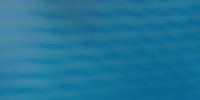}}\hspace{4pt}
		\subfloat{\label{}\includegraphics[width=0.15\linewidth]{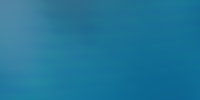}}\hspace{4pt}
		\subfloat{\label{}\includegraphics[width=0.15\linewidth]{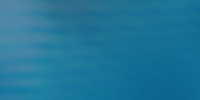}}\hspace{4pt}
		\subfloat{\label{}\includegraphics[width=0.15\linewidth]{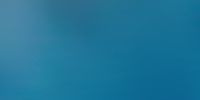}}\hspace{4pt}
	\end{minipage}
	\begin{minipage}[ht]{.99\linewidth}
		\vspace{-7pt}
		\centering
		\subfloat[\fontsize{8pt}{12pt} \selectfont Focused]{\label{}\includegraphics[width=0.15\linewidth]{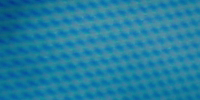}}\hspace{4pt}
		\subfloat[\fontsize{8pt}{12pt} \selectfont Defocused]{\label{}\includegraphics[width=0.15\linewidth]{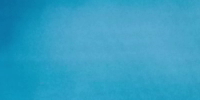}}\hspace{4pt}
		\subfloat[\fontsize{8pt}{12pt} \selectfont MMDM]{\label{}\includegraphics[width=0.15\linewidth]{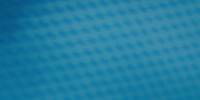}}\hspace{4pt}
		\subfloat[\fontsize{8pt}{12pt} \selectfont BPN]{\label{}\includegraphics[width=0.15\linewidth]{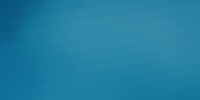}}\hspace{4pt}
		\subfloat[\fontsize{8pt}{12pt} \selectfont VDM$_{PCD}$]{\label{}\includegraphics[width=0.15\linewidth]{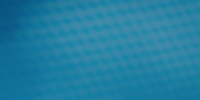}}\hspace{4pt}
		\subfloat[\fontsize{8pt}{12pt} \selectfont Ours]{\label{}\includegraphics[width=0.15\linewidth]{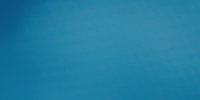}}\hspace{4pt}
	\end{minipage}
	\centering
	\fontsize{8pt}{12pt} \selectfont (d) Ablation of network\\
        \vspace{-5pt}
	\caption{Ablation study in the four aspects.}
\label{fig:ablation}
\end{figure*}

\begin{figure*}
\includegraphics[width=\linewidth]{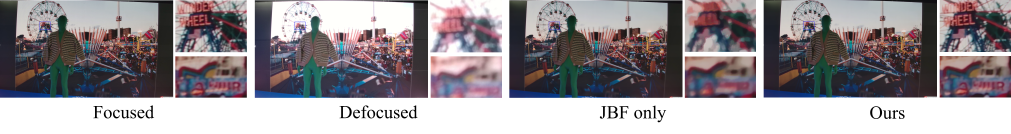}
\caption{Ablation study of 'JBF only'.}
\label{fig:JBF_only_1}
\end{figure*}

\begin{figure*}
\includegraphics[width=\linewidth]{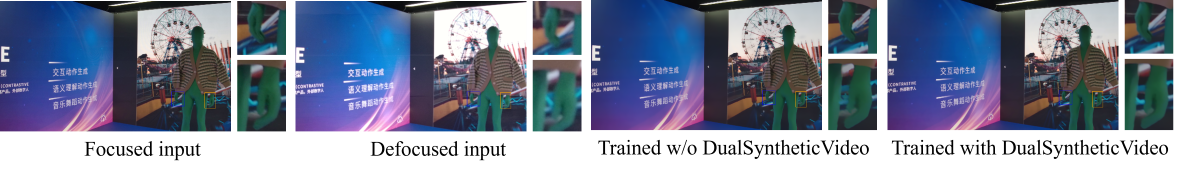}
\caption{Ablation study of training our method without and with the training data of DualSyntheticVideo.}
\label{fig:withDualSyntheticVideo}
\end{figure*}

\subsection{Dataset and experimental details}
We create three focused-defocused dual-camera demoireing datasets, namely the real dataset (DualReal), the synthetic image dataset (DualSynthetic) and the synthetic video dataset (DualSyntheticVideo).

To build DualReal, We use a real dual-camera setup (Fig. \ref{fig:inconsistency}) to collect 50 pairs of focused-defocused videos. The resolution is $1920 \times 1080$. We display images and videos on the LCD screen and use the dual-camera to shoot towards the screen to obtain the videos. We use the shutter release cable to ensure synchronization of the two cameras. We use a tripod to fix the two cameras and put them as close as possible, as shown in the dual camera of Fig. \ref{fig:inconsistency}, to reduce occlusion area between the pair of focused and defocused videos. Because we perform optical flow instead of stereo matching in our framework, we do not require the traditional camera calibration using the checkerboard. The displayed images are from 5K dataset \cite{else_67} and the displayed videos are from some popular movies. We also generate a synthetic image dataset, i.e. DualSynthetic, and a synthetic video dataset, i.e. DualSyntheticVideo, as introduced in Sec. \ref{sec:synthetic}. In DualSynthetic, randomly divided 4700/1000 pairs are used for training/testing. In DualSyntheticVideo, randomly divided 12/11 videos are used for training/testing.

In training the demoireing models, we mix the training data of DualSyntheticVideo with the training data of DualSynthetic. DualSynthetic contains plenty of images and types of moire patterns due to the random generation pipeline, and input pairs of frames in DualSyntheticVideo contain misalignment and occlusions which are more close to the real data captured by dual-camera systems. The trained models are tested on DualReal.

All experiments are performed on NVIDIA RTX 3090 GPU. We follow \cite{moire_2_27} \cite{moire_28} \cite{moire_4_34} to use the metrics of PSNR, SSIM and LPIPS for image quality measurement. We also use the metrics of t-MSE and temporal SSIM (t-SSIM) to measure the temporal consistency quality of videos.

In the alignment step of our method, we use the Flowformer \cite{10.1007/978-3-031-19790-1_40} model finetuned on the Sintel dataset\cite{sintel}. In estimating the optical flow, to reduce the computational costs, we downsample the input frame from $1920 \times 1080$ to $704 \times 396$ with Bilinear interpolation and feed the downsampled frames to the model. The estimated flow is resized and rescaled to the original size by Bilinear interpolation too. Our demoireing network is implemented by PyTorch. The model uses Adam \cite{else_71} as the optimizer. During the training process, we crop the image as $256 \times 256$, set the batch size to 1, and the number of model iterations as 100. The learning rate is initially set to 0.0002.  The joint bilateral filter has three parameters, \textit{i.e.}  window size, range sigma and spatial sigma, and we set them as 51, 10, and 10 in our experiments.

\subsection{Comparison with State-of-the-Arts}
We compare our method with the state-of-the-arts methods on the datasets DualReal, DualSynthetic and DualSyntheticVideo we created. The comparison algorithms include the single-image demoireing methods of DMCNN \cite{moire_2_27}, MopNet \cite{moire_28}, MBCNN \cite{moire_31}, WDNet \cite{moire_4_34}, ESDNet \cite{moire_6_36}, FDNet \cite{moire_11_40}, the multi-frame demoireing algorithm MMDM \cite{moire_41}, the video demoireing algorithm VDM$_{PCD}$ \cite{moire_14_42}, the general image restoration methods of Uformer \cite{transformer_43},  SwinIR \cite{transformer_45}, and the general multi-image restoration methods of SiamTrans \cite{transformer_44_13}, MFD\cite{restoration_54}. The single-image demoireing methods and general restoration methods were retrained on DualSynthetic dataset. For single-frame algorithms, we extract each frame in sequence from the focused video and process them one by one. For multi-frame algorithms, we use multiple temporally adjacent frames as references for the current frame.

As the results show in Figs. \ref{fig:result_1}-\ref{fig:sintel_result_1} and Tables \ref{tabel:contrast synthetic}-\ref{tabel:contrast real}, since all comparison methods assume that, during the test stage, the input video/image comes from a single camera, they cannot effectively utilize the defocused frames from the secondary camera. These methods struggle to distinguish between moire patterns and textures, which are often similar in real scenarios. Consequently, their demoireing results either fail to effectively remove moire patterns or introduce artifacts and color inconsistencies.

The defocused frames provide valuable guidance to differentiate between moire patterns and textures in the contents of the focused frames. Leveraging the defocused frames from the secondary camera of our dual-camera setup, our proposed pipeline achieves improved moire pattern removal while preserving textures. Additionally, in the recovery step, due to the thoughtful framework design and the indirect use of the result from the demoireing network to generate the final output, our results also successfully attain the goals of tonal and temporal consistency.

\begin{table*}[h]
\caption{ \centering Average PSNR, SSIM and LPIPS values on DualSynthetic.}
\renewcommand\arraystretch{0.3}
\tabcolsep=2.5pt
\small
%\vspace{-10pt}
\begin{center}
\begin{tabular}{cccccccccccccc}
\toprule
 & DMCNN & MopNet & MBCNN & WDNet & Uformer & ESDNet & SwinIR & MFD & MMDM & SiamTrans & VDM$_{PCD}$ & FDNet & \quad Ours\\
\midrule
 PSNR & 28.56 & 29.47 & 24.80 & 26.54 & 29.56 & 29.74 & 28.15 & 28.47 & 22.38 & 23.78 & \underline{33.38} & 26.44 & \quad \textbf{34.38}\\
 SSIM  & 0.8166 & 0.8538 & 0.6227 & 0.6624 & 0.8336 & 0.8485 & 0.7971 & 0.8277 & 0.4839 & 0.6871 & \underline{0.8908} & 0.6717 & \quad \textbf{0.9151}\\
 LPIPS  & 0.2482 & 0.1705 & 0.3963 & 0.2128 & 0.2022 & 0.1579 & 0.2785 & 0.2431 & 0.4943 & 0.3842 & \underline{0.1510} & 0.3481 & \quad \textbf{0.1243}\\
\bottomrule
\end{tabular}
\end{center}
\label{tabel:contrast synthetic}
\end{table*}

\begin{table*}[h]
\caption{ \centering Average PSNR, SSIM, LPIPS, t-MSE and t-SSIM values on DualSyntheticVideo.}
\renewcommand\arraystretch{0.3}
\tabcolsep=2.5pt
\small
%\vspace{-10pt}
\begin{center}
\begin{tabular}{cccccccccccccc}
\toprule
 & DMCNN & MopNet & MBCNN & WDNet & Uformer & ESDNet & SwinIR & MFD & MMDM & SiamTrans & VDM$_{PCD}$ & FDNet & \quad Ours\\
\midrule
 PSNR & \underline{33.88} & 29.91 & 30.62 & 31.03 & 32.86 & 32.25 & 33.48 & 33.35 & 29.93 & 26.31 & 32.63 & 28.69 & \quad \textbf{39.23}\\
 SSIM  & \underline{0.8972} & 0.7801 & 0.7621 & 0.7775 & 0.8650 & 0.8446 & 0.8794 & 0.8944 & 0.7046 & 0.8223 & 0.8547 & 0.8010 & \quad \textbf{0.9750}\\
 LPIPS  & 0.0569 & 0.1044 & 0.1145 & 0.0723 & 0.0551 & 0.0658 & 0.0734 & 0.0567 & 0.1819 & 0.1353 & \underline{0.0549} & 0.0970  & \quad \textbf{0.0137}\\
  t-MSE  & 52.48 & 63.04 & 66.52 & 64.11 & 56.91 & 57.85 & 53.46 & 52.23 & 70.92 & \textbf{44.46} & 56.27 & 57.89 & \quad \underline{44.89}\\
    t-SSIM  & 0.6712 & 0.5288 & 0.4928 & 0.5265 & 0.6166 & 0.6037 & 0.6623 & 0.6722 & 0.4432 & \textbf{0.8104} & 0.6142 & 0.5971 & \quad \underline{0.7439} \\
\bottomrule
\end{tabular}
\end{center}
\label{tabel:contrast sintel}
\end{table*}

\begin{table*}[h]
\caption{ \centering Average t-MSE and t-SSIM values on DualReal.}
\renewcommand\arraystretch{0.3}
\tabcolsep=2.5pt
\small
%\vspace{-10pt}
\begin{center}
\begin{tabular}{cccccccccccccc}
\toprule
 & DMCNN & MopNet & MBCNN & WDNet & Uformer & ESDNet & SwinIR & MFD & MMDM & SiamTrans & VDM$_{PCD}$ & FDNet & \quad Ours\\
\midrule
  t-MSE  & 23.73& 23.72 & 25.95 & 30.45 & 25.35 & 26.52 & 23.09 & 24.16 & 26.83 & \underline{22.81} & 24.31 & 22.83 & \quad \textbf{21.84}\\
    t-SSIM  & 0.8881 & 0.8789 & 0.8602 & 0.8349 & 0.8678 & 0.8756 & \underline{0.9070} & 0.8841 & 0.8511 & \textbf{0.9119} & 0.8873 & 0.8994 & \quad 0.8955 \\
\bottomrule
\end{tabular}
\end{center}
\label{tabel:contrast real}
\end{table*}

\subsection{Ablation Study}

We test different variants of our method in four aspects, and the results are shown in Fig. \ref{fig:ablation}. \textbf{Training Data:} The variants include that (1) we do not add the foreground to synthetic images, named `No foreground'. As shown in Fig. \ref{fig:ablation} (a), in `No foreground', the model will over-smooth the foreground objects. \textbf{Training Loss:} The variants of only using some of the multi-dimensional losses (consistency, perceptual, high-frequency, and paired adversarial loss) are tested, but any variant will have a decrease in demoireing. This verifies that each of the four losses is beneficial to obtain higher quality results. \textbf{Framework:} The variants include that (1) we only use the single focused frame as the input without using the defocused frame as guidance, named `No defocus', (2) the joint bilateral filter in Eq. \ref{eqn:JBF} is not performed and $\bf{I_R}$ is directly used as the output, named `No JBF', and (3) we don't warp the defocused frame to align with the focus frame, named `No alignment'. In `No defocus', it is difficult for the model to determine moire from texture and thus generate poor results. In `No JBF', the moire patterns in occlusion regions can hardly be removed completely, and the inconsistency of brightness and color with the input focused frame cannot be solved. In `No alignment', ghosting artifacts may appear. \textbf{Network:} (1) we replace the demoireing network with the MMDM network \cite{moire_41}, named `MMDM', (2) the BPN network \cite{restoration_53}, named 'BPN', and (3) the VDM$_{PCD}$ network \cite{moire_14_42}, named `VDM$_{PCD}$'. This verifies the advantage of the adopted network in this paper.

We also take an ablation study to use $\bf{I_A}$ as guide to perform joint bilateral filter for $\bf{I_F}$ to obtain the demoireing result. This ablation study is named ‘JBF only’ and the results are shown in Fig. \ref{fig:JBF_only_1}. As shown, there are many lost details in the aligned defocused image $\bf{I_A}$ due to the defocused setting. When ‘JBF only’ uses $\bf{I_A}$ as guide, the guide image cannot provide accurate guidance for the detailed textures during the joint bilateral filtering process, leading to the lost of details in the filtering results. 

In addition, we take an ablation study to train our method using DualSynthetic (named without DualSyntheticVideo) or using the mix of DualSynthetic and DualSyntheticVideo (named with DualSyntheticVideo). The testing results on DualReal are shown in \ref{fig:withDualSyntheticVideo}. As shown, with the help DualSyntheticVideo to train our method, the occlusion regions of the testing results have improved quality, because the input pairs of frames in DualSyntheticVideo contain misalignment and occlusions and can encourage the demoire network to learn to deal with these regions during training.

\section{Conclusion}
The paper introduces a focused-defocused dual-camera system for video demoireing. The proposed processing framework aims to remove moire patterns while preserving the texture, maintain tonal consistency, and avoid temporal flickering. The frame-wise processing framework incorporates contextual information between the input focused and defocused frames, and consists of an alignment step to solve displacement and occlusion between the frames, a demoireing step that uses a multi-scale CNN and a multi-dimensional training loss to remove moire patterns, and a recovery step to address tonal and temporal inconsistencies. The experimental results demonstrate the superiority of our method over state-of-the-art demoireing methods.

\section{Acknowledgment}
We thank Junjie Zhong for his help with the experiment during the manuscript revision process. We thank Tianfan Xue for the insightful discussion during the manuscript revision process. This work was supported in part by the National Key R \& D Program of China under Grant 2022ZD0161901, BUPT Kunpeng \& Ascend Center of Cultivation, the Fundamental Research Funds for the Central Universities, and the Beijing Nova Program under Grant 20230484297.

\bibliographystyle{IEEEtran}
\bibliography{egbib}

\vspace{-1.5cm}
\begin{IEEEbiographynophoto}{Xuan Dong}
received the Ph.D. degree in Computer Science from Tsinghua University in 2015, and the B.E. degree in Computer Science from Beihang University in 2010. He is currently an Associate Professor with Beijing University of Posts and Telecommunications, China. His research interests include computer vision and computational photography.
\end{IEEEbiographynophoto}

\vspace{-1.5cm}
\begin{IEEEbiographynophoto}{Xiangyuan Sun}
is currently pursuing the M.E. degree in Artificial Intelligence at Beijing University of Posts and Telecommunications. His current research interests include computer vision and image processing.
\end{IEEEbiographynophoto}

\vspace{-1.5cm}
\begin{IEEEbiographynophoto}{Xia Wang}
is currently pursuing the M.E. degree in Artificial Intelligence at Beijing University of Posts and Telecommunications. Her current research interests include computer vision and image processing.
\end{IEEEbiographynophoto}

\vspace{-1.5cm}
\begin{IEEEbiographynophoto}{Jian Song}
is the CTO of DeepScience Co., Ltd. DeepScience is a company to provide professional video service, including the building of the shooting scene, video shooting systems, and video streaming service.
\end{IEEEbiographynophoto}

\vspace{-1.5cm}
\begin{IEEEbiographynophoto}{Ya Li}
received the B.E. from the University of science and technology of China, in 2007, and the PhD degree from the Institute of automation, Chinese Academy of Sciences, in 2012. She is currently an Associate Professor with the Beijing University of Posts and Telecommunications. Her research interests include speech synthesis, and multimodal interaction. 
\end{IEEEbiographynophoto}

\vspace{-1.5cm}
\begin{IEEEbiographynophoto}{Weixin Li}
received the Ph.D. degree in computer science from the University of California at Los Angeles (UCLA), Los Angeles, CA, USA, in 2017. She is currently an Associate Professor at the School of Computer Science and Engineering, Beihang University, Beijing, China. Her research interests include computer vision, image processing, and big data analytics.
\end{IEEEbiographynophoto}

\end{document}